\documentclass[runningheads]{llncs}

\usepackage{eccv}

\usepackage{eccvabbrv}

\usepackage{graphicx}
\usepackage{booktabs}
\usepackage{makecell}
\usepackage{multirow}
\usepackage{subcaption}

\usepackage[accsupp]{axessibility}  %

\usepackage[textsize=tiny]{todonotes}
\usepackage{ifthen}
\newboolean{comments}
\setboolean{comments}{false} %

\usepackage{hyperref}
\usepackage[capitalize]{cleveref}

\usepackage{orcidlink}

\usepackage{algorithm}
\usepackage{multicol}
\usepackage{float}          %
\usepackage{algpseudocode}  %
\usepackage{amsmath}
\usepackage[table]{xcolor} %
\usepackage{wrapfig}

\usepackage[most]{tcolorbox}
\newtcolorbox{promptbox}[2][]{%
    enhanced,
    breakable,
  colback=gray!5,
  colframe=gray!75,
  title=#2,
  fonttitle=\bfseries,
  #1
}
\newcommand{\tagheading}[1]{%
  \par\smallskip
  \noindent\textbf{#1}\par
}
\tcbuselibrary{breakable}
\usepackage{enumitem}
\usepackage{longtable}

\definecolor{headerblue}{RGB}{31, 78, 121}  %
\definecolor{rowblue}{RGB}{230, 242, 255}   %
\definecolor{LightRed}{RGB}{255,230,230}
\definecolor{DarkRed}{RGB}{255,180,180}
\definecolor{LightGreen}{RGB}{220,255,220}
\definecolor{paleblue}{RGB}{235,245,255}

\newcommand{\EVE}{\texttt{EVE}\xspace}

\begin{document}

\title{EVE: A Generator-Verifier System for Generative Policies} 

\titlerunning{\EVE: A Generator-Verifier System for Generative Policies}

\author{Yusuf Ali$^{*}$ \inst{1} \and
Gryphon Patlin$^{*}$ \inst{1} \and Karthik Kothuri$^{\dagger}$\inst{1} \and Jeremiah Coholich$^{\dagger}$\inst{1} \and Muhammad Zubair Irshad\inst{2} \and Wuwei Liang\inst{3} \and Zsolt Kira\inst{1}}

\makeatletter
\renewcommand{\footnoterule}{%
  \kern -3pt
  \hrule \@width 0.48\textwidth
  \kern 2.6pt
}
\makeatother

\begingroup
\renewcommand{\thefootnote}{}
\footnotetext{\textsuperscript{*}Co-first authors. \textsuperscript{$\dagger$}Co-second authors.}
\endgroup

\authorrunning{Ali et al.}

\institute{Georgia Institute of Technology \and
Toyota Research Institute \and
Symbotic Inc.}

\maketitle

\begin{abstract}

Visuomotor policies based on generative such as diffusion and flow-matching have shown strong performance for robotics applications but degrade under distribution shifts,  demonstrating limited recovery capabilities without costly finetuning.
In the language modeling domain, test-time compute scaling has revolutionized the reasoning capabilities of modern LLMs by enabling candidate solution refinement.
These methods typically leverage foundation models as verification modules in a zero-shot manner to score candidate solutions.
We hypothesize that generative policies can similarly benefit from additional inference-time compute that employs zero-shot VLM-based verifiers in a generation-verification framework.
To this end, we introduce \EVE: a modular, generator-verifier interaction framework that boosts the performance of pretrained generative policies at test time, with no additional training.
\EVE wraps a frozen base policy with multiple zero-shot, VLM-based verifier agents.
Each verifier proposes action refinements to the base policy candidate actions, while an action incorporator uses classifier guidance to fuse aggregated verifier feedback into action denoising.
We study design choices for generator–verifier information interfacing across a system of verifiers with distinct capabilities. 
Across diverse simulated and real robotic tasks and embodiments, \EVE consistently improves success rates without additional policy or verifier training.
Through extensive ablations, we isolate the contribution of verifier capabilities and action incorporator strategies, offering practical guidelines to build scalable, modular generator-verifier systems for embodied control.

\end{abstract}
    
\section{Introduction}
\label{sec:intro}

Foundation models for embodied tasks have demonstrated strong performance across a variety of complex tasks. 
These Vision Language-Action (VLA) models are typically trained on a large set of manually collected robot demonstrations \cite{bjorck2025gr00t,pizero,kim2024openvla,lbmpaper,team2025gemini} through imitation learning.
VLAs generate action outputs with diffusion \cite{liu2024rdt} or flow-matching \cite{pizero} architectures which provide high inference speed and ability to model complex and multimodal continuous action distributions \cite{urain2024deep}.
Although these models have shown promising application in diverse robot manipulation task setups, they struggle with slight deviations to operating conditions (such as tabletop heights) and do not exhibit strong recovery capabilities when encountering out-of-distribution states during deployment \cite{gu2025safe,zhai2025vision,yang2025fpc,black2025real}.
Improving the performance or robustness of such generative policies is typically done by finetuning or retraining with additional in-domain data (or recovery sequences), which is expensive to collect \cite{dai2025racer,lin2025failsafe,yang2025fpc}.
Furthermore, the performance of finetuned policies is significantly affected by such finetuning routines and is heavily dependent on the scale and quality of the finetuning data \cite{robocasa365,yadav2026robust}.

In lieu of finetuning, we propose to leverage frontier vision-language models (VLMs) within a unified generator-verifier architecture, which we term \EVE (Embodied Verifier Ensembles). 
Scaling test-time compute by leveraging learned reward models (or \textit{verifiers}) has fundamentally redefined the capabilities of foundation LLMs without any additional finetuning or retraining \cite{cobbe2021training,lightman2023let,uesato2022solving,snell2024scaling,zhang2024generative}.
These works typically sample multiple candidate solutions from the base LLM and score them using external verifiers to maximize downstream task performance.
In this work, we argue that a similar shift is underway in the embodied domain, wherein frozen, pretrained generative policies can be improved using similar \textit{zero-shot verifiers} at test-time deployment.
Additionally, this paradigm is well-suited to the robotics domain as collecting high-quality, real-world data is an expensive and laborious routine \cite{o2024open,khazatsky2024droid,bu2025agibot}.
While recent work has begun leveraging such verifiers for downstream embodied tasks,
these approaches require training the verifier module or latent dynamics models \textit{tabula rasa} \cite{kwok2025robomonkey,wu2025foresight}. 
In contrast, \EVE orchestrates multiple zero-shot, VLM-based verifier agents that are focused on distinct capabilities that boost the performance of the frozen base generator policy.
Through extensive experimentation, we systematically study various design choices that affect the interaction between the generator policy and verifier modules to improve task performance on a diverse set of embodied manipulation tasks.

There are several key challenges in implementing a generator-verifier architecture for visuomotor policies. (1) Unlike in language modeling, where both the generator and verifier LLM operate in a shared modality (vision or text), embodied settings involve continuous robot actions, which complicates information interfacing (i.e. how should action information be communicated between the generator and verifier); 
(2) Most LLM generator-verifier systems use best-of-N sampling routines, where multiple candidate outputs are generated and the verifier selects the best one based on the downstream task — but in the embodied domain, selection over multiple continuous action trajectories for task completion
is non-trivial. The issue is exacerbated if verifiers provide feedback in distinct ways (e.g, only selecting trajectories or choosing from a set of predefined primitives)
(3) In embodied tasks, it is unclear how to combine the verifier’s aggregated action output with the base policy action predictions as naive action averaging or verifier action overrides might not be optimal.

To overcome these challenges, we propose querying multiple verifiers with distinct action-feedback strategies and combining the output action corrections with the base policy through a classifier-guidance based action incorporator.

Specifically, the key contributions of our work are as follows:
\begin{enumerate}
    \item We propose \texttt{EVE}, a \textbf{generator-verifier} system tailored for embodied policies in which verifiers operate with different input modalities, capabilities, and action spaces to improve test-time policy performance. 
    \item We propose an \textbf{action incorporator} module that employs guided diffusion to fuse aggregated verifier outputs with action predictions from base policy.
   \item We show that zero-shot \EVE ensembles \textbf{outperform state-of-the-art embodied verifier baselines} trained with substantial in-domain data on \textit{SimplerEnv} ($7$ diverse tasks across $2$ embodiments), and further extend to long-horizon mobile manipulation tasks on \textit{Maniskill-HAB}, and tabletop manipulation tasks in \textit{RoboTwin} and a real-world Franka manipulator.

    \item We further present analysis on how \EVE's ensemble \textbf{recovers failed rollouts} and show how \textbf{verifier ensembles} outperform individual counterparts. We conduct extensive ablations and analyses that provide a detailed understanding of the design choices in building effective generator-verifier systems.

\end{enumerate}

\section{Related Work}
\label{sec:related_work}

\textbf{Test-Time Scaling Through Verification.} Recent work has found that spending additional compute during test-time deployment of LLMs can lead to large gains in performance on complex reasoning tasks \cite{wei2022chain,snell2024scaling}. 
Specifically, many recent works have focused on leveraging the generation-verification gap wherein additional LLMs are employed to \textit{verify} the output of the base generator LLM through learned outcome \cite{cobbe2021training} or process-based reward models \cite{lightman2023let}.
\cite{zhou2025variation} conducts a systematic study of the interaction dynamics between generator and verifier models for text-only reasoning problems.
There have been few recent works which have started leveraging the generation-verification framework for improving embodied task performance.  
Robomonkey \cite{kwok2025robomonkey} proposes to train a reward model from scratch using synthetically-mined action preferences from a large-scale robotics dataset.
The learned verifier is then used to score action predictions from the base VLA policy.
HAVE \cite{li2025learn} proposes to train a history-conditioned verifier which is used to score outputs from a diffusion-based generator policy to achieve reduced failure rates during task execution.
In contrast to these works, we instead propose to build a system of \textit{zero-shot} verifiers which can be used to boost the robustness of the base generative policy.
MAV \cite{lifshitz2025multi} proposes a system of heterogeneous verifiers that work in conjunction to verify different aspects of the candidate solutions generated by the base LLM.
They constrain their study to text-only LLM verifiers and mathematical/factual questions.
In \texttt{EVE}, we construct an ensemble of zero-shot verifiers with \textit{distinct capabilities} and develop a systematic policy-verifier orchestration through an action incorporator module that interpolates between policy prediction and verifier feedback.

\textbf{Reasoning/Steering to Improve Embodied Policies.} There has been tremendous progress in improving performance of LLMs by scaling test-time compute using additional token generation \cite{wei2022chain}.
Recent work has focused on training policies with reasoning capabilities to improve generalization of the policy in diverse task settings \cite{zawalski2024robotic,chen2025training,clark2025action}. 
These works typically require fully finetuning the large VLA policy on reasoning traces
but do not possess the ability to leverage additional compute during inference.
Another line of work focuses on \textit{steering} the behavior of embodied policies towards desired objectives during task execution. 
These works typically focus on learning a latent dynamics model which is used to simulate future states to compute alignment with the task goal state.
The misalignment with the desired task completion state is then used to compute an error signal which is used in classifier-guidance style steering \cite{du2025dynaguide,sun2025latent,wang2025inference} or simpler post-hoc ranking of corresponding action proposals \cite{nakamoto2024steering,wu2025foresight}. 
In contrast to explicit steering, SAILOR \cite{jain2025smooth} tries to discover new recovery sequences within a learned world model and distill them into the base policy using imitation learning.
They additionally learn a reward model which is used to score latent states obtained from the world model.
In contrast to the aforementioned works which require policy training from scratch or learning latent world models, we propose a method to improve policy performance through a \textit{generator-verification} framework that comprises multiple \textit{zero-shot verifiers}.

\vspace{-10pt}
\section{Methodology}
\label{sec:methodology}

\begin{figure*}[t]
    \centering
    \includegraphics[width=1.0\linewidth]{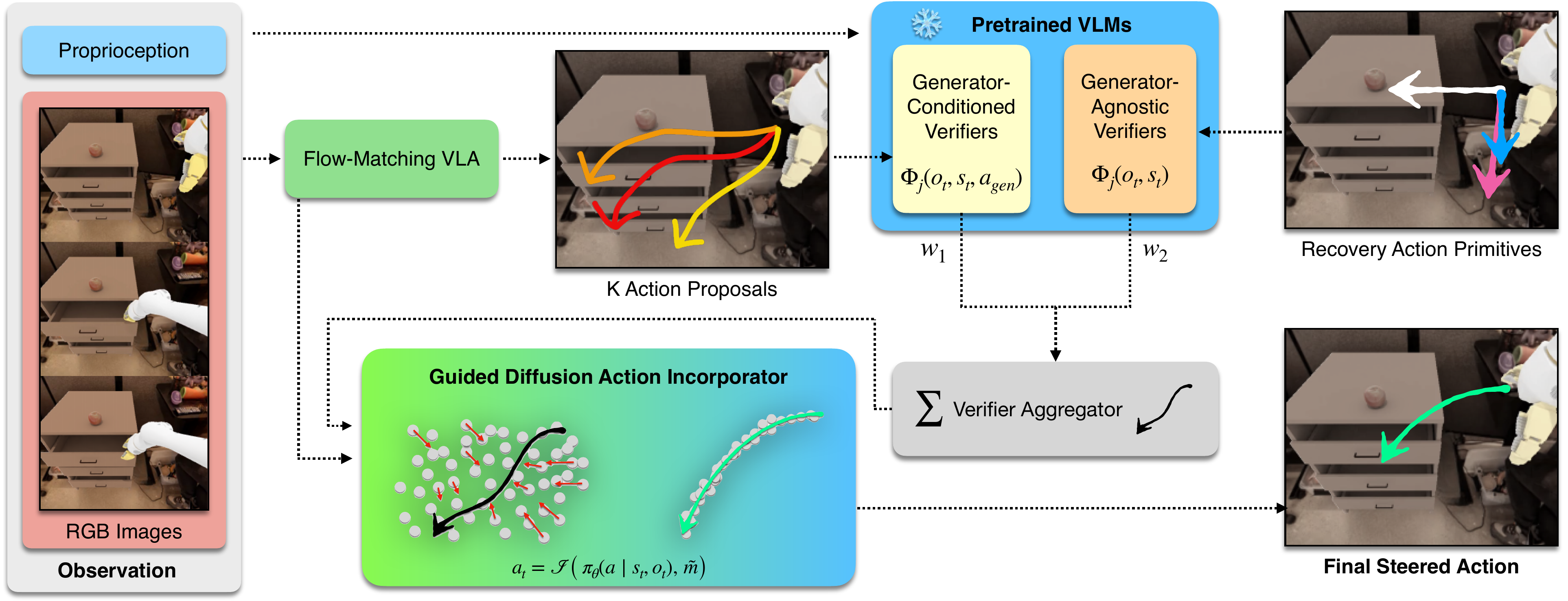}
    \caption{\EVE: A Generator-Verifier Interaction Framework for Generative Embodied Policies. Action feedback from ensemble of verifier agents is incorporated with base policy denoising through diffusion guidance.}
    \label{fig:placeholder}
    \vspace{-1mm}
\end{figure*}

We propose \texttt{EVE} a unified framework that augments pre-trained generative policies with modular verifier agents to improve action quality through multi-agent output aggregation and action incorporation.

\subsection{Base Policy Candidate Generation}

At each timestep $t$, given an instruction $x$, observation $o_t$ (e.g., RGB or depth), proprioceptive state $s_t$ (e.g., end-effector joint positions), and a frozen base policy $\pi_\theta$, we generate a set of candidate actions:
\begin{small}
\begin{equation}
    a_{\text{gen}} = \left\{ a_t^{(k)} \right\}_{k=1}^{K},
    \label{eq:candidate_actions}
\end{equation}
\end{small}
where each candidate action $a_t^{(k)}$ is produced by the base policy as
\begin{small}
\begin{equation}
    a_t^{(k)} \sim \pi_\theta(o_t, s_t).
    \label{eq:base_policy_action}
\end{equation}
\end{small}
\looseness=-1 In the case of a frozen diffusion policy, these candidate actions correspond to action sequences that are denoised from $K$ independent noise samples, representing diverse plausible trajectories conditioned on the current observation and state.

\subsection{Verifier Agents}
\label{sec:verifier_agents}

We define a collection of \textbf{Verifier Modules} $\mathcal{V} = \{ V_j \}_{j=1}^J$, each endowed with a specific capability to improve the base policy action generation. 
Each verifier follows a contract:
\begin{small}
\begin{equation}
V_j:\; \Phi_j(o_t, s_t, a_{\text{gen}}) \rightarrow m_j \in \mathcal{M}_j
\end{equation}
\end{small}
Here, $\Phi_j$ denotes a verifier-specific encoding of input context and candidate actions, and $m_j(a_{\text{gen}})$ is a message in a structured output space $\mathcal{M}_j$ such as trajectory selections or text-based action corrections.
We note that a subset of the verifiers can also operate without access to the base policy action proposals. 
We propose to categorize the verifiers in our framework based on the information available from the generator policy and the specific capability that each verifier focuses on.
We provide a detailed categorization as follows:

\textbf{Generator-Agnostic:} 
These verifiers operate primarily on the observations from the robot sensor and do not leverage action information from the generator policy.
Specifically, this category of verifiers constitute the entire set of $V_j$ for which the assiciated policy-verifier interface $\Phi_j(o_t, s_t, a_{\text{gen}})$ always recevies $a_{\text{gen}} = \varnothing$
These verifiers are conditioned only on the task instruction and are required to select action sequences that maximally ensure task completion.

\textbf{Generator-Conditioned:} 
These verifiers take action information from the generator policy as inputs to provide action feedback.
For instance, such a verifier takes a representation of the candidate action sequences as input - in addition to raw observations from the robot sensors.
In this work, we consider trajectory-based representations used in prior work \cite{nasiriany2024pivot} to relay information between the base policy and verifier but methods leveraging alternate representations are equally applicable in our framework \cite{liu2024moka,huang2023voxposer}.
In essence, the above verifier categorization implicitly defines a \textit{generator-verifier interface} wherein each verifier module $V_j$ interacts with the generator policy through the encoding $\Phi_j$ defined for that specific verifier.

\textbf{Verifier Output Aggregation.} 
We first bring the output of each verifier $m_j$ into an action trajectory representation.
When using a \textit{generator-agnostic} verifier we directly ask the verifier module to select from a list of predefined action primitives, each of which has a corresponding trajectory sequence. 
In contrast, \textit{generator-conditioned} verifiers directly select action trajectories from the set of available base policy actions (see App \cref{app:verfier_ensemble} for details).
We then introduce an aggregation operator $\mathcal{A}$ that projects individual verifier outputs $m_j$ into a common semantic space for unified information relay back to base policy.
Formally, given the set of verifier outputs $\{m_j\}_{j=1}^J$, the aggregated verifier output is defined as
\begin{small}
\begin{equation}
  \tilde{m} = \mathcal{A}\big(\{\,m_j)\,\}_{j=1}^J\big),
\end{equation}
\end{small}

\noindent where $\tilde{m}$ denotes the fused trajectory resulting from a weighted interpolation across verifier outputs.
In ~\Cref{sec:ablations}, we ablate the effects of different weighting strategies in $\mathcal{A}$ (based on verifier type) on downstream task success rates.

\subsection{Action Incorporator}
\label{sec:guided_diffusion}

We define an \textit{action incorporator} that fuses the base policy output with the aggregated verifier trajectory $\tilde{m}$.  
At each intervention step $t$, the executed action is
\begin{small}
\begin{equation}
  a_t = \mathcal{I}\big(\,\pi_\theta(a \mid s_t, o_t),\; \tilde{m}\big),
\end{equation}
\end{small}

\noindent where $\pi_\theta(a \mid s_t, o_t)$ is the base policy action conditioned on the input context.
The action incorporator $\mathcal{I}$ leverages a mechanism that reconciles the base policy output with the verifier-derived trajectory signal $\tilde{m}$ to generate the final executable control $a_t$.
Instead of directly overriding or averaging with the base policy action, the incorporator progressively refines the action through a guided denoising process that steers the policy toward verifier-consistent behavior while preserving the original policy prior.

In the Guided Diffusion (GD) framework, action synthesis is directed by an objective function $\xi(\tau, z)$, which encodes the alignment between the generated trajectory $\tau$ and a verifier-derived feedback signal $z$.
At each diffusion timestep $k$, given an observation-state sample $(o_t, s_t)$ and the noisy action sample $a_t^k$, the reverse diffusion step is expressed as:
\begin{small}
\begin{equation}
a_{t}^{k-1} = \alpha_k \Big( a_t^k - \gamma_k \big( \epsilon_\theta(o_t, a_t^k, k) + \beta_k \nabla_{a_t^k} \xi(a_t^k, z) \big) \Big) + \sigma_k \eta,
\label{eq:guided_action_diffusion}
\end{equation}
\end{small}

\noindent where $\epsilon_\theta(o_t, a_t^k, k)$ denotes the denoising network conditioned on the current observation $o_t$, state $s_t$, and diffusion step $k$, and $\eta \sim \mathcal{N}(0, I)$ represents Gaussian noise.
The diffusion-specific hyperparameters $\alpha_k$, $\gamma_k$, and $\sigma_k$ are derived from the DDPM noise scheduler (with a squared cosine $\beta$-schedule), which defines the forward–reverse diffusion dynamics and noise variance at each timestep.
Most importantly, the guidance coefficient $\beta_k$ controls the influence of the alignment gradient derived from the verifier feedback.

The alignment gradient $\nabla_{a_t^k} \xi(a_t^k, z)$ is computed using action-level feedback from a verifier system. 
Specifically, given an action trajectory $z$ which represents the verifier system preferences, we define the objective function as the L2-norm discrepancy between the generated action and the verifier feedback:
\begin{small}
\begin{equation}
    \xi(a_t^k, z) = \frac{1}{2} \| a_t^k - z \|_2^2.
    \label{eq:alignment_objective_action}
\end{equation}
\end{small}
The corresponding gradient is therefore:
\begin{small}
\begin{equation}
    \nabla_{a_t^k} \xi(a_t^k, z) = a_t^k - z,
    \label{eq:alignment_gradient_action}
\end{equation}
\end{small}

\noindent which provides a simple and effective alignment direction that minimizes the action discrepancy with respect to verifier feedback. This gradient term biases the reverse diffusion process toward generating verifier-consistent actions while maintaining stability within the learned conditional distribution $p(a_t \mid o_t,s_t)$ of the pretrained policy. 
This ensures that the final denoised actions are coherent with verifier-approved behaviors.

\vspace{-1mm}
\subsection{Intervention Detection}
\label{sec:intervention_detection}

We note that computing verifier feedback at each step in the rollout can be an expensive routine since it requires multiple VLM inference calls.
To counter this, we propose to invoke verifier feedback only at specific intervention points which are automatically detected as the rollout progresses.
We leverage an off-the-shelf failure detector for generative policies \cite{agia2024unpacking} that uses statistical measures to flag erratic failures during action execution.
Using the aforementioned, the \texttt{EVE} system is invoked whenever the cumulative maximum mean discrepancy (MMD) in the trajectory exceeds a threshold value.
We briefly review the cumulative MMD computation in the following.

For some offset $k$, let $\bar{\pi}_t := \pi(a_{t+k:t+h-1} \mid s_t)$ and $\tilde{\pi}_{t+k} := \pi(a_{t+k:t+h-1} \mid s_{t+k})$ denote the marginal action distributions over the temporally overlapping action segments between timesteps $t$ and $t + k$. We define the temporal consistency between two contiguous timesteps $t$ and $t+k$ as $ \hat{D}(\bar{\pi}_t, \tilde{\pi}_{t+k}) \ge 0$, where $\hat{D}$ represents MMD metric. In our case, we set our offset $k$ to be the execution horizon of the given policy. (see App \cref{app:mmd_detail} for details).

\subsection{Putting it all together: \texttt{EVE}}
\label{sec:eve_end_to_end}

We bring together all the individual components discussed in the preceeding sections to build an inference-time action refinement algorithm.
We highlight that our proposed generator-verifier framework generates semantically grounded action feedback from VLM-based verifiers and seamlessly interpolates it with the base policy action distribution through a guided diffusion framework.
We require no additional finetuning of the policy weights to induce recovery.
We outline the pseudocode in \cref{alg:eve}.

\begin{algorithm}
\scriptsize
\caption{\EVE: Embodied Verifier Ensemble Inference Pseudocode}
\label{alg:eve}

\newcommand{\codecomment}[1]{\textcolor{blue!60}{\texttt{// #1}}}

\begin{algorithmic}[1]

\Require Horizon $H$, Observations $\{o_t\}$, states $\{s_t\}$, Frozen base policy $\pi_\theta$ with $N$ denoising steps, Verifiers $\mathcal{V}=\{V_j\}_{j=1}^J$, MMD threshold $\tau$, MMD computation samples $M$

\For{$t = 1$ \textbf{to} $H$}

    \State \codecomment{Candidate Action Generation}
    \State Sample $K$ base-policy candidates $a_{\mathrm{gen}}=\{a_t^{(k)}\}_{k=1}^K$ using $\pi_\theta$
    \Comment{Eqs.~(\ref{eq:candidate_actions}), (\ref{eq:base_policy_action})}

    \State Compute MMD score $\eta_t$ from overlapping segments using $K$ action samples
    \Comment{Eq.~(\ref{eq:mmd_equation})}
    \If{$\eta_t < \tau$}
        \State Execute nominal action from $\pi_\theta$ and \textbf{continue}
    \EndIf

    \State \codecomment{Verifier Inference}
    \For{$j = 1$ \textbf{to} $J$}
        \State Build verifier-specific encoding $\Phi_j(o_t,s_t,a_{\text{gen}})$
        \State $m_j \gets V_j(\Phi_j(o_t,s_t,a_{\text{gen}}))$
        \Comment{Selections/corrections over $a_{\text{gen}}$}
    \EndFor
    \State $\tilde{m} \gets \mathcal{A}(\{m_j\}_{j=1}^J)$
    \Comment{Aggregate verifier outputs}

    \State \codecomment{Guided Diffusion Action Incorporation}
    \State Initialize noisy action sample $a_t^{N}$ (from DDPM prior)

    \For{$k = N, \ldots, 1$}

        \State Set $z \gets \tilde{m}$ and define
        \State \quad $\xi(a_t^k,z) = \tfrac{1}{2}\lVert a_t^k - z\rVert_2^2$
        \Comment{Eq.~(\ref{eq:alignment_objective_action})}

        \State Compute alignment gradient
        \State \quad $g_k \gets a_t^k - z$
        \Comment{Eq.~(\ref{eq:alignment_gradient_action})}

        \State Denoise with guidance:
        \State \quad
        $a_t^{k-1} \gets
        \alpha_k \bigl(a_t^k - \gamma_k(\epsilon_\theta(o_t,a_t^k,k) + \beta_k g_k)\bigr)
        + \sigma_k \eta$
        \Comment{Eq.~(\ref{eq:guided_action_diffusion})}

    \EndFor

    \State $a_t \gets a_t^{0}$
    \Comment{Final executable control}
    \State Execute $a_t$

\EndFor

\end{algorithmic}

\end{algorithm}

\section{Implementation Details}
\label{sec:implementation_details}

\textbf{\texttt{EVE} Verifier Details.} 
For the \textit{Generator-Conditioned} verifier, we employ the PIVOT \cite{nasiriany2024pivot} prompting strategy where we supply $40$ samples from the base policy.
We then draw on to the observation image $5$ most visually distinct trajectories based on the cosine similarity metric.
We refer to this as \textit{Pivot} steerer in all experiments.
For the \textit{Generator-Agnostic} verifier, we mark the goal location of the target object on the RGB image at the intervention point and ask the VLM to suggest a recovery action from a set of action primitives. 
We only utilize camera views that are accessible to the base policy in all benchmark settings (see App \cref{app:task_setup} for details).
We refer to this as \textit{Primitive} steerer in all experiments.
We provide all prompts used in our experiments in App \cref{app:prompts} and details on verifiers in App \cref{app:verfier_ensemble}.
We use vLLM \cite{kwon2023efficient} as the primary inference engine in our experiments (see App. \cref{app:vllm_details} for details).
For all simulation experiments, we use the \texttt{Qwen-2.5-VL-72B} \cite{bai2025qwen2} as the backbone VLM for all verifiers.

\textbf{Benchmark and Evaluation Setup.} 
We presents results on the SimplerEnv benchmark \cite{simplerenv} which is a real-to-sim evaluation benchmark for tabletop tasks that has high correlation with real-world success rates.
We conduct evaluations on $7$ tasks across two embodiments: $4$ from WidowX arm and $3$ from the Google Robot.
On SimplerEnv, we use $\pi_0$ \cite{pizero} as the base policy to show that \EVE can improve the performance of large VLA policies (see \cref{app:pi_zero_res}).
We report results on $48$ rollouts across $3$ random seeds.
In addition we present results on the open-source \textit{ManiSkill-HAB} long-horizon, mobile manipulation benchmark \cite{shukla2024maniskill} which provides various tasks that require precise contact-rich manipulation for successful task completion.
We also evaluate \EVE on the \textit{Robotwin-2.0} \cite{robotwin2} task suite that employs a bimanual arm to establish generalization to more complex embodiments. 
We provide further details on the base policy and task setup in App. \cref{app:task_setup}.
We provide detailed description of baselines and evaluation protocol in \cref{app:baseline_evals}.
Detailed hyperparameter setting are provided in \cref{app:hyperparameters}.

\section{Experiments}
\label{sec:experiments}

\looseness=-1  We focus on the following research questions in this work: 
1) Do ensembles of zero-shot verifiers outperform state-of-the-art embodied verifiers trained with in-domain data ? (\cref{sec:main_result}), 2) Can VLM-based zero-shot verifiers in \EVE generalize to new embodiments and longer-horizon tasks? (\cref{sec:long_horizon_tasks}), 3) How does verifier-based steering with \EVE qualitatively alter failure trajectories during execution? (see \cref{sec:qual_analysis}), 4) How do various individual components in \texttt{EVE} affect task performance? (\cref{sec:ablations}), and 5) How does \EVE manage latency? (\cref{sec:latency_analysis}) 6) Can \EVE generalize to real-world tasks (\cref{sec:real_robot_experiments})

\section{Results}

\subsection{Performance on SimplerEnv}
\label{sec:main_result}

\begin{table}[t]
\centering
\setlength{\tabcolsep}{2pt}
\renewcommand{\arraystretch}{1.0}
\begin{tabular}{lc|c|c|ccc}
\toprule
\textbf{Task} &
$\pi_0$ & V-GPS & RoboMonkey &
\multicolumn{3}{c}{\EVE} \\
& \scriptsize{\cite{pizero}}
& \scriptsize{\cite{nakamoto2024steering}}
& \scriptsize{\cite{kwok2025robomonkey}}
& \scriptsize{Pivot} & \scriptsize{Primitive} & \scriptsize{Ensemble} \\

\cmidrule(lr){1-4} \cmidrule(lr){5-7}
\multicolumn{1}{l}{\textbf{Verifier Training Budget}}
& -
& \cellcolor{LightRed}\scriptsize 175K demos
& \cellcolor{DarkRed}\scriptsize 20M synthetic
& \multicolumn{3}{c}{\cellcolor{LightGreen}\scriptsize \textbf{0 (zero-shot)}} \\

\midrule
\multicolumn{7}{l}{\textbf{WidowX}} \\
\midrule
Carrot on Plate & $56.2$ & $13.2$ & $54.9$ & $\underline{56.9}$ & $56.2$ & $\mathbf{60.4}$ \\
Eggplant Basket & $91.7$ & $20.1$ & $88.2$ & $91.0$ & $86.1$ & $\mathbf{94.4}$ \\
Spoon on Towel & $84.7$ & $2.1$ & $83.3$ & $\underline{87.5}$ & $\mathbf{88.9}$ & $\mathbf{88.9}$ \\
Stack Blocks & $56.2$ & $4.2$ & $58.3$ & $\mathbf{63.2}$ & $\underline{61.1}$ & $\underline{61.1}$ \\
\midrule
Average & $72.2$ & $9.9$ & $71.2$ & $74.7$ & $73.1$ & $\mathbf{76.2}$ \\

\midrule
\multicolumn{7}{l}{\textbf{Google Robot}} \\
\midrule
Move Near & $77.1$ & $77.8$ & $\underline{80.6}$ & $\underline{80.6}$ & $79.9 $ & $\mathbf{84.7}$ \\
Put Apple in Drawer & $29.9$ & $4.9$ & $35.4$ & $\underline{36.1}$ & $33.3$ & $\mathbf{40.3}$ \\
Close Drawer & $73.6$ & $68.8$ & $\mathbf{75.7}$ & $\underline{73.6}$ & $70.8$ & $\mathbf{75.7}$ \\
\midrule
Average & $60.2$ & $50.5$ & $63.9$ & $63.4$ & $61.3$ & $\mathbf{66.9}$ \\

\midrule
Total Average & $67.1$ & $27.3$ & $68.1$ & $69.8$ & $68.0$ & $\mathbf{72.2}$ \\
\bottomrule
\end{tabular}
\vspace{1mm}
\captionof{table}{\EVE with training-free, \textbf{zero-shot verifiers} outperforms state-of-the-art finetuned verifiers on the SimplerEnv benchmark \cite{simplerenv}. Verifier training budgets are shown beneath method names. \textbf{Bold} is best performing and \underline{underline} is second-best.}
\label{tab:simplerenv_results}
\vspace{-20pt}
\end{table}

We present the main quantitative results in \cref{tab:simplerenv_results}, evaluating \EVE on SimplerEnv \cite{simplerenv} across a diverse set of manipulation tasks on two different embodiments and analyze performance gains relative to prior approaches that leverage \emph{trained} verifiers for policy steering.

\textbf{\EVE achieves the highest performance across diverse tasks and embodiments.}
Across both robot embodiments, \EVE improves over the base policy $\pi_0$ and prior verifier-based baselines.
In particular, \texttt{EVE-Ensemble} attains the best total average success rate of ${72.2}$, outperforming the base $\pi_0$ ($67.1$) and improving upon steering gains through trained verifiers such as RoboMonkey ($68.1$) and V-GPS ($27.3$).
\texttt{EVE-Ensemble} achieves $\mathbf{76.2}$ average success on WidowX and $\mathbf{66.9}$ on Google Robot, indicating robust performance across different platforms and tasks.

\looseness=-1 \textbf{\EVE outperforms trained verifier baselines.}
From \cref{tab:simplerenv_results} we see that \EVE outperforms recent competitive verifier baselines that require substantial in-domain training budgets.
For example, V-GPS \cite{nakamoto2024steering} is trained with $175$K demonstrations, while RoboMonkey \cite{kwok2025robomonkey} relies on $20$M synthetic action preference samples for verifier training.
More importantly, these verifiers are trained on task and embodiment-specific data which leads to limited generalization capabilities.
In contrast, \EVE uses zero training data, operating entirely with zero-shot VLM-based verifiers.
Despite this dramatic reduction in data requirements, \EVE variants achieve higher average performance across the task suite.
We hypothesize that these benefits partially stem from \EVE's capability to provide verification feedback at the \textit{action chunk level}.
In contrast, the trained verifiers score individual actions on a per-step basis which can lead to poor temporal coherence.

\textbf{Ensembling Improves Individual Verifiers.}
Even without ensembling, \texttt{EVE-Pivot} achieves $69.8\%$ and \texttt{EVE-Primitive} achieves $68.0\%$ on the total average, indicating that both verifier configurations effectively improve the base policy through our proposed generator-verifier framework.
However, ensembling these zero-shot verifiers yields the strongest overall performance, suggesting complementary feedback from each verifier that better captures task progress and impending failures.

\textbf{Task-wise large improvements are concentrated on harder, lower success tasks.}
Analyzing task-wise performance of \texttt{EVE-Ensemble}, the largest jumps occur on lower-performing tasks, such as \emph{Put Apple} in Drawer ($29.9 \rightarrow 40.3$) and \emph{Stack Blocks} ($56.2 \rightarrow 63.2$).
We also observe strong gains on \emph{Move Near} (77.1 $\rightarrow$ 84.7) and \emph{Carrot on Plate} (56.2 $\rightarrow$ 60.4).
However, improvements are modest on near-saturated tasks such as \emph{Eggplant Basket} (91.7 $\rightarrow$ 94.4), and on \emph{Close Drawer}, we find that \texttt{EVE-Ensemble} matches the best performance, suggesting limited remaining headroom.

\subsection{Performance on Maniskill-HAB and RoboTwin}
\label{sec:long_horizon_tasks}

\begin{table}[t]
\centering
\small

\begin{subtable}{0.48\linewidth}
\centering
\resizebox{\linewidth}{!}{
\begin{tabular}{lccc}
\toprule
\textbf{Task} & DP & DP + \EVE & $\Delta$ \\
\midrule
SetTable-OpenFridge      & 64.09 & \textbf{66.27} & +2.18 \\
SetTable-Place           & 56.85 & \textbf{58.63} & +1.78 \\
SetTable-Pick            & 25.40 & \textbf{26.19} & +0.79 \\
PrepareGroceries-Place   & 35.02 & \textbf{35.22} & +0.20 \\
PrepareGroceries-Pick    & \textbf{11.70} & 11.28 & -0.42 \\
TidyHouse-Pick           & 16.07 & \textbf{16.17} & +0.10 \\
\bottomrule
\end{tabular}
}
\phantomcaption
\label{tab:mshab_results}
\caption*{(a)}
\vspace{-6pt}
\end{subtable}
\hfill
\begin{subtable}{0.48\linewidth}
\centering
\resizebox{\linewidth}{!}{
\begin{tabular}{lccc}
\toprule
\textbf{Task} & $\mathbf{\pi_{0.5}}$ & $\mathbf{\pi_{0.5}}$ + \textbf{\EVE} & $\Delta$\\
\midrule
Beat Block Hammer & $41.67\%$ & $\mathbf{45.49\%}$ & $+3.82\%$\\
Place Can Basket & $18.40\%$ & $\mathbf{21.88\%}$ & $+3.48\%$ \\
Place Container Plate & $80.56\%$ & $\mathbf{83.68\%}$ & $+3.12\%$\\
Place Object Stand & $48.61\%$ & $\mathbf{49.31\%}$ & $+0.7\%$\\
Move Can Pot & $\mathbf{20.83\%}$ & $20.14\%$ & $-0.69\%$\\
\bottomrule
\end{tabular}
}
\phantomcaption
\label{tab:robotwin_performance}
\caption*{(b)}
\vspace{-6pt}
\end{subtable}

\caption{(a) \EVE consistently steers a base diffusion policy towards higher success on mobile manipulation tasks in Maniskill-Hab \cite{shukla2024maniskill}. (b) \texttt{EVE} improves performance of $\mathbf{\pi_{0.5}}$ VLA \cite{pi05} the RoboTwin 2.0 benchmark \cite{robotwin2}. \textbf{Note:} RoboMonkey \cite{kwok2025robomonkey} and V-GPS \cite{nakamoto2024steering} are not applicable since they require extensive in-domain training to operate on new tasks and embodiments.}
\label{tab:combined_tables}
\end{table}

\textbf{Maniskill-HAB Results.} In this section, we apply \EVE to long-horizon mobile manipulation tasks from the Maniskill-HAB benchmark \cite{shukla2024maniskill}.
We present results in \cref{tab:mshab_results}.
We note that other state-of-the-art verifiers, such as RoboMonkey \cite{kwok2025robomonkey} and V-GPS \cite{nakamoto2024steering}, are not applicable since they are \textbf{not trained} to provide verification feedback for complex mobile manipulation tasks on new embodiments.
We report $\Delta \text{SR} = \text{SR}_{\text{steered}} - \text{SR}_{\text{unsteered}}$ and employ the \texttt{EVE-Ensemble} configuration.
We report the mean $\Delta \text{SR}$ over $1008$ rollouts to ensure statistical significance.
From \cref{tab:mshab_results}, we observe that \EVE delivers consistent improvements in performance above the unsteered base policy rollouts by leveraging additional zero-shot verifiers.
We note that the external verifiers are \textit{not finetuned} with any task-specific data and are able to provide benefits in diverse task settings and environments.
Specifically, we observe the largest benefits in \texttt{SetTable-OpenFridge} and \texttt{SetTable-Place} with improvements of $2.18\%$ and $1.78\%$ respectively.
This empirically proves that \EVE is able to boost the performance of the diffusion policy on complex tasks that require recovering from subtle execution degradation.
We also present a detailed ablation study on MSHAB in \cref{sec:ablations}.

\textbf{Robotwin-$2.0$ Results.} To establish further generalization capability of \EVE, we provide extended results on a new task suite and bimanual arm embodiment.
For these experiments, we evaluate using the Aloha AgileX dual-arm embodiment.
\cref{tab:robotwin_performance} shows that \EVE steering leads to performance gains over the $\mathbf{\pi_{0.5}}$ VLA on RoboTwin tasks.
We note that this is a particularly strong result since other verifier baselines such as RoboMonkey \cite{kwok2025robomonkey} and V-GPS \cite{nakamoto2024steering} are not applicable in a zero-shot way to Robotwin tasks as they require significant in-domain training. 
We apply diffusion guidance to $\mathbf{\pi_{0.5}}$ following the same technique described in \cref{app:pi_zero_res}.

\subsection{A Closer Look: How \EVE  recovers failed trajectories?}
\label{sec:qual_analysis}

In this section, we perform a qualitative analysis of $2$ different scenarios in which \EVE steers a failed episode towards a successful completion.

\begin{figure}[!h]
    \centering

    \begin{subfigure}{\linewidth}
        \centering
        \includegraphics[width=\linewidth,height=0.25\textheight,keepaspectratio]{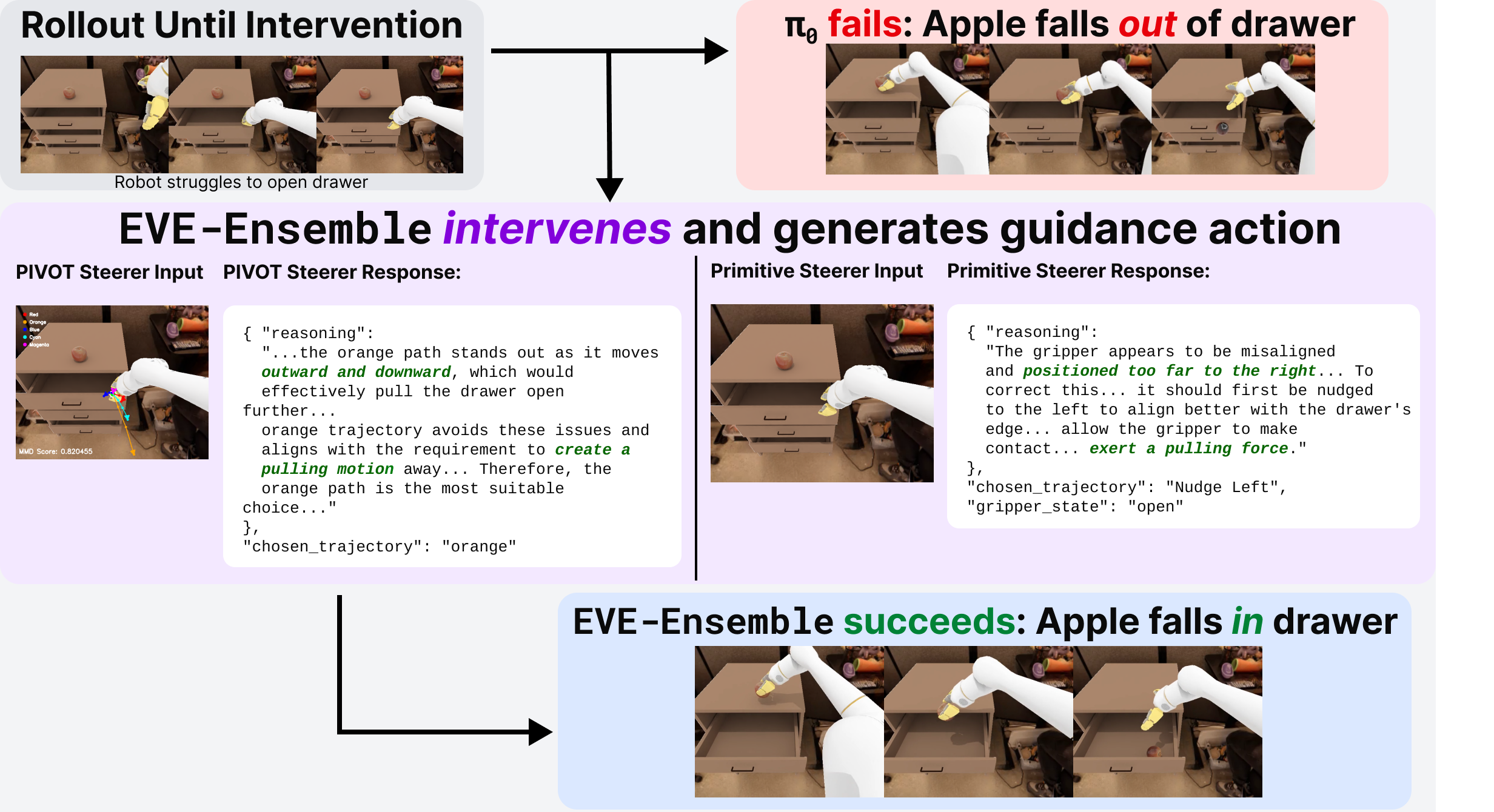}
        \caption{\textbf{Complementary strengths.} The verifiers provide complementary guidance that helps recover a failed trajectory on \texttt{Place Apple} task failed trajectory.}
        \label{fig:place_apple_a}
    \end{subfigure}

    \vspace{1em}

    \begin{subfigure}{\linewidth}
        \centering
        \includegraphics[width=\linewidth,height=0.25\textheight,keepaspectratio]{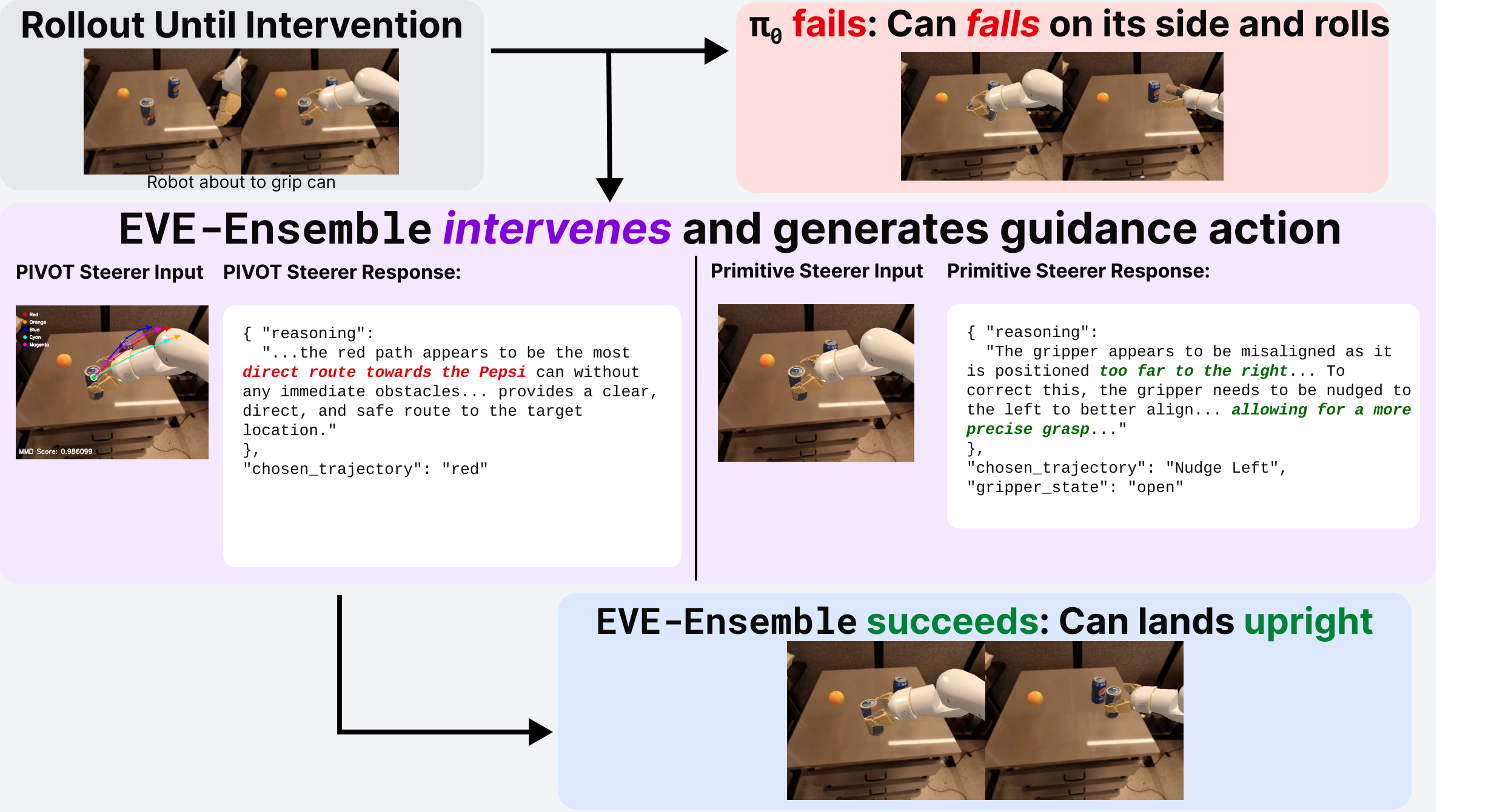}
        \caption{\textbf{Robustness via ensembling.} Aggregating verifier feedback remains corrective even when one steerer is misleading at an intervention point during failed rollout of \texttt{Move Near} task.}
        \label{fig:move_near}
    \end{subfigure}

    \caption{\EVE steering case studies on SimplerEnv.}
    \label{fig:place_apple}

\vspace{-5pt}
\end{figure}

\textbf{Complementary verifier capabilities recover failed trajectory.}
In this case, we analyse a failed rollout of the task \texttt{Place Apple in Drawer}, a task that requires the robot to open the drawer and place the apple inside it.
In this particular rollout, the base policy $\pi_0$ is only able to open the drawer partially even after repeated attempts. 
Consequently, when it tries to place the apple inside the drawer, it leads to misaligned placement, causing the apple to fall out rather than being placed successfully as shown in \cref{fig:place_apple_a}.
In contrast, \EVE detects a large spike in the MMD value of action samples during drawer opening and intervenes to guide the policy towards success.
At the intervention point, the image input is sent to the ensemble where the \textit{Primitive} steerer proposes a “Nudge Left” action to guide the gripper toward the handle, while the \textit{Pivot} steerer selects trajectories (from the proposed base trajectories) that \textit{maximize outward pulling }of the drawer and discard misaligned ones.
The averaged feedback from both the verifier actions is used to guide the final action denoising.
This intervention enables the drawer to open fully and results in the successful placement of the apple, and qualitatively shows how the complementary guidance from two different verifiers helps recover a failed trajectory.

\textbf{Robustness in verifier intervention via Ensembling}.
We analyze another failed rollout on the \texttt{Move Near} task, which requires the robot to grasp the Red Bull can and place it upright near the Coke can.
As shown in \cref{fig:move_near}, the base policy $\pi_0$ attempts to grasp the can but executes a faulty grip, causing the can to tip over and roll off the table.
In contrast, \EVE detects a spike in the MMD value during the \textit{approach phase} and triggers an intervention.
At the intervention point, the \textit{Primitive} steerer proposes a corrective \textit{Nudge Left} action to better align the gripper with the can, improving the grasp configuration.
Interestingly, the \textit{Pivot} steerer incorrectly selects a trajectory that does not correspond to meaningful task progress.
However, because \EVE aggregates feedback across verifiers, averaging the ensemble feedback yields a small leftward correction that leads to achieving a stable grasp. 
This enables the robot to grasp the Red Bull can correctly and place it upright near the Coke can.
Overall, this example illustrates \EVE's robustness to imperfect verifier feedback: even if one steerer is misleading at an intervention point, the aggregated ensemble signal remains corrective and recovers the trajectory.

\subsection{Ablating Individual Components within \texttt{EVE}}
\label{sec:ablations}

\begin{figure*}[htbp]
\centering

\resizebox{0.65\textwidth}{!}{%
\begin{minipage}{\textwidth}
\centering
\begin{subfigure}[t]{0.30\textwidth}
  \centering
  \includegraphics[width=\linewidth]{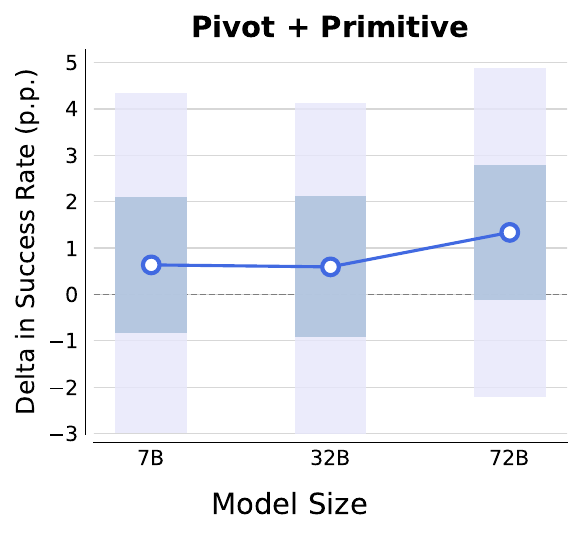}
  \caption{}
  \label{fig:model_scaling}
\end{subfigure}
\hfill
\begin{subfigure}[t]{0.30\textwidth}
  \centering
  \includegraphics[width=\linewidth]{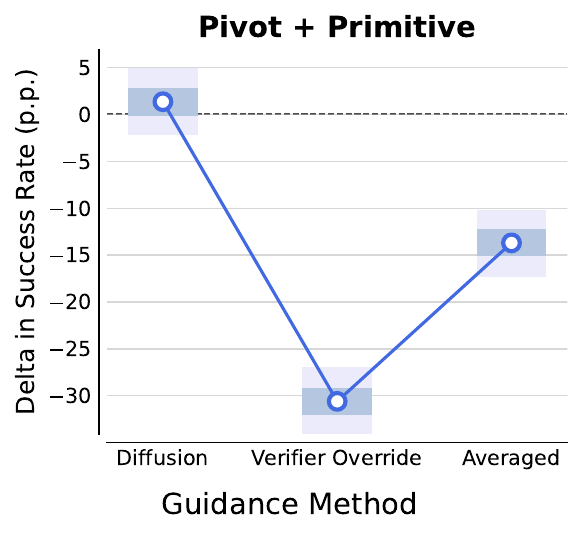}
  \caption{}
  \label{fig:guide_strategy}
\end{subfigure}
\hfill
\begin{subfigure}[t]{0.30\textwidth}
  \centering
  \includegraphics[width=\linewidth]{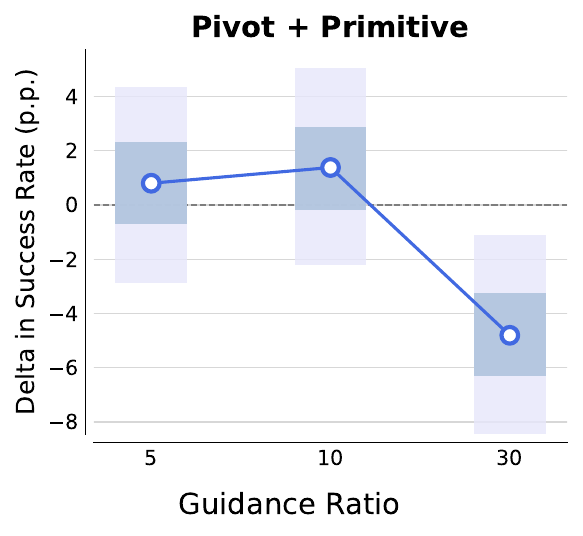}
  \caption{}
  \label{fig:guide}
\end{subfigure}
\end{minipage}
}

\vspace{-2mm}
\caption{\EVE ablations on the MSHAB task suite (system-level components).}
\label{fig:eve_ablation_system}
\vspace{-10pt}
\end{figure*}

In this section we present extensive ablations of each module in the \EVE framework.
For these experiments, we use the \texttt{SetTable-Place} task from MSHAB  using the \texttt{EVE-Ensemble} configuration, unless otherwise stated.
In all ablation results, we report the \textit{Delta in Success Rate \%} which measures the delta gain in steered and unsteered runs.
See \cref{sec:detailed_ablations} for ablations on verifier inputs.

\textbf{Verifier Model Scaling.}
We investigate how the size of the underlying VLM affects the verifier's ability to steer the base policy effectively.
We compare the performance impact using Qwen-2.5-VL at different parameter scales: $7$B, $32$B, and $72$B.
As illustrated in \cref{fig:model_scaling}, increasing the parameter scale of the verifier model size generally correlates with improved task success rates. 
Specifically, we find that the $72$B model consistently outperforms the smaller variants.
This finding is in-line with recent work from the language modelling literature \cite{zhang2024generative,zhou2025variation} which find increasing benefits with stronger verifier models.

\textbf{Action Incorporator Design.}
In \cref{fig:guide_strategy}, we ablate various strategies to incorporate verifier-aggregated action feedback into the base policy action predictions.
From the results, it is clear that the \textit{Verifier Override} leads to a drastic reduction in performance. 
This is potentially because the \textit{Primitive} verifier only selects from a predefined list of recovery primitives which are used for guidance over the base action dimensions only (see App \cref{app:verfier_ensemble} for details).
Additionally, we observe that direct averaging between verifier-aggregated and base policy actions performs poorly in comparison to the guided diffusion strategy employed in \EVE. This is because direct averaging does not ensure that the output action is close to the marginal action distribution of the base policy.
This suggests that the action incorporator in \EVE integrates \emph{''just''} the right amount of verifier feedback to prevent task failure but still ensure task completion.

\textbf{Guidance Ratio Ablation.}
We conduct ablations with diffusion guidance coefficient $\beta_k$ (see \cref{eq:guided_action_diffusion}), which controls amount of verifier feedback incorporated into base policy denoising.
In \cref{fig:guide}, we observe that guidance coefficient significantly affects performance of the task with an optimal value of $10$ with sharp drops in neighboring values.
This result suggests that a very large value of guidance pushes the denoising too far away from the base policy action distribution, causing large temporal inconsistencies leading to reduced task performance.

\subsection{Latency Analysis}
\label{sec:latency_analysis}

\begin{wraptable}{r}{0.5\textwidth}
\vspace{-8mm}
\centering
\setlength{\tabcolsep}{1pt}
\renewcommand{\arraystretch}{1.0}
\resizebox{\linewidth}{!}{%
\begin{tabular}{@{}lcccc@{}}
\toprule
\textbf{Task} & $\pi_0$ & RoboMonkey & \multicolumn{2}{c}{\texttt{EVE-Ensemble}} \\
\cmidrule(lr){4-5}
  & -- & --
 & \scriptsize{\shortstack{\texttt{Qwen3-VL}\\\texttt{8B}}}
 & \scriptsize{\shortstack{\texttt{Qwen2.5-VL}\\\texttt{72B}}} \\
\midrule
\multicolumn{5}{@{}l}{\textbf{Success Rate $\uparrow$}} \\
\textit{WidowX} Avg.
  & 72.2
  & 71.2
  & \underline{75.2}
  & \textbf{76.2} \\

\textit{Google Robot} Avg.
  & 60.2
  & 63.9
  & \underline{64.8}
  & \textbf{66.9} \\
Overall
  & 67.1
  & 68.1
  & \underline{70.7}
  & \textbf{72.2} \\
\midrule
\multicolumn{5}{@{}l}{\textbf{Latency $\downarrow$}} \\
Per Step (s) 
  & --
  & \textbf{1.01}
  & \underline{3.59}
  & 11.59 \\
Avg. Rollout Time (s)
  & --
  & 378.0
  & \underline{357.8}
  & \textbf{333.4} \\
\bottomrule
\end{tabular}%
}
\vspace{-3mm}
\caption{\EVE with smaller \textit{but newer} VLM verifier backbones outperforms RoboMonkey on SimplerEnv. Although \EVE has higher per-step latency, the average rollout time is lower than RoboMonkey.}
\label{tab:efficiency_comparison}
\vspace{-6mm}
\end{wraptable}

From \cref{tab:efficiency_comparison}, we first observe that \EVE equipped with the smaller and newer \texttt{Qwen3-VL-8B} backbone still outperforms RoboMonkey, which is finetuned with $20$M synthetic action preferences, suggesting that the performance gains from verification with \EVE are likely to scale with stronger and newer VLM backbones.
We further observe that \EVE achieves a comparable average rollout time to RoboMonkey.
This is primarily because RoboMonkey finetunes the LLaVA-7B model to score candidate actions at \textit{every timestep} of the rollout.
Instead \EVE leverages a Maximum Mean Discrepancy (MMD) trigger to selectively invoke the VLM-based verifier only when the distribution of sampled actions from the base policy deviates significantly (see \cref{sec:intervention_detection}), and generates verification feedback at the \textit{action chunk level}.
This leads to a much lower intervention frequency, resulting in comparable average rollout timing.
We note, however, that \EVE with the smaller \texttt{Qwen3-VL-8B} backbone still has higher \textit{per-step} verification latency than RoboMonkey, since the latter uses only a single VLM call without token generation.
We hope that future stronger and smaller verifiers can reduce this per-step latency.
We also provide a latency and throughput analysis on the MSHAB task suite in \cref{app:verifier_overhead}.

\subsection{Real Robot Experiments}
\label{sec:real_robot_experiments}

\begin{figure}[t]
    \centering

    \begin{subfigure}{0.48\linewidth}
        \centering
        \includegraphics[width=\linewidth]{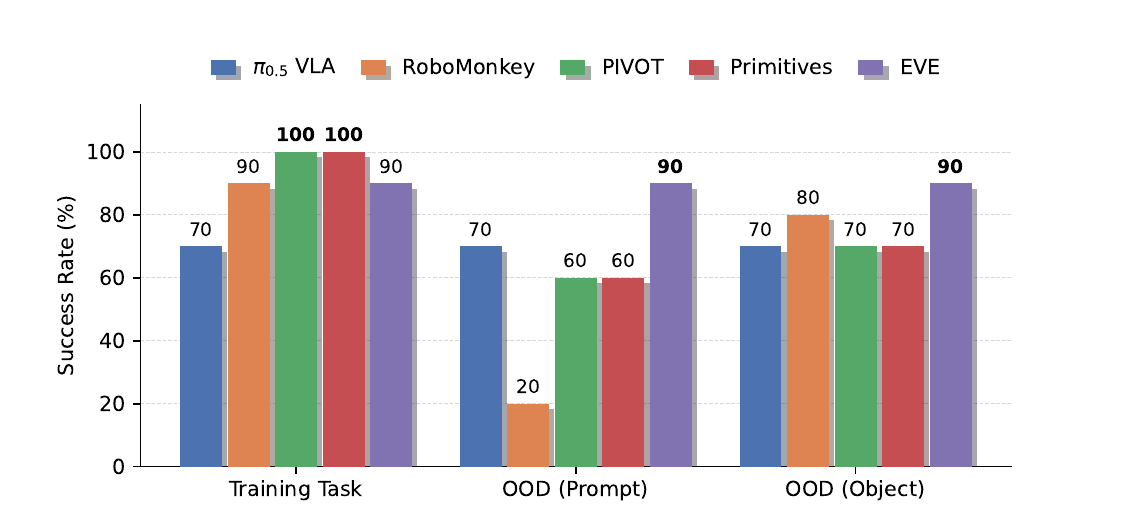}
        \caption{}
        \label{fig:real-world-SR}
    \end{subfigure}
    \hfill
    \begin{subfigure}{0.5\linewidth}
        \centering
        \includegraphics[width=\linewidth]{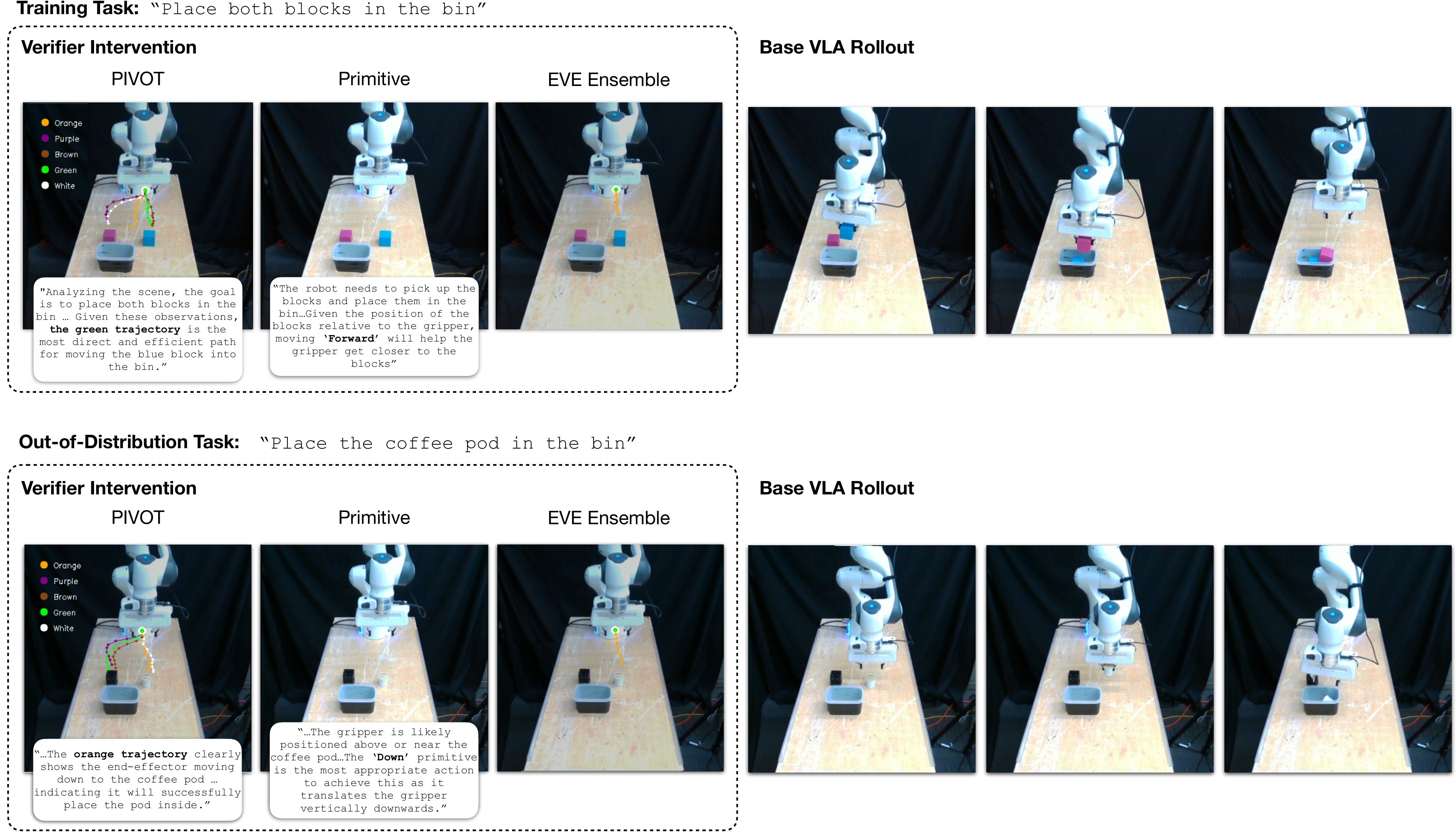}
        \caption{}
        \label{fig:real_world}
    \end{subfigure}

    \caption{(a) Each bar represents a success rate out of $10$ trials. \textit{OOD-Prompt} requires picking blocks in a specific order and \textit{OOD-Object} requires manipulating unseen object. \EVE performs best in OOD task settings on real Franka robot. (b) Sample real-world rollouts demonstrate how \EVE provides recovery feedback. \textit{``Base VLA Rollout"} denotes rollout continuation \textit{after} \EVE \textit{intervention}.}
    \label{fig:real_world_combined}
\end{figure}

We validate \EVE with real-world robot tabletop manipulation experiments. 
We collect $50$ demonstrations using a Franka Emika Panda arm to place two blocks into a bin.
The layout of objects can be seen in \cref{fig:real_world}.
We finetune the $\pi_{0.5}$ VLA \cite{pi05} with an action chunk size of $100$ and we query \EVE for the first $2$ action chunks but RoboMonkey is queried at each chunk.
\cref{fig:real-world-SR} shows success rates on real robot evaluations using \EVE on $5$ tasks: $1$ in-distribution task (\textit{place both blocks in bin}); $2$ OOD-Prompt tasks (pick in specific order) and $1$ OOD-Object task (pick coffee pod).
\EVE improves performance of $\pi_{0.5}$ on ID tasks and matches RoboMonkey.
Across $3$ OOD tasks, \EVE \textit{achieves the best performance} by leveraging semantically-informed verifier-guided corrections.
RoboMonkey degrades in the \textit{OOD Prompt} setting showing limited zero-shot applicability.

\subsection{Further Analysis and Limitations}
\label{sec:failures}

\begin{wrapfigure}{r}{0.35\textwidth}
\vspace{-20pt}
    \centering
    \includegraphics[width=\linewidth]{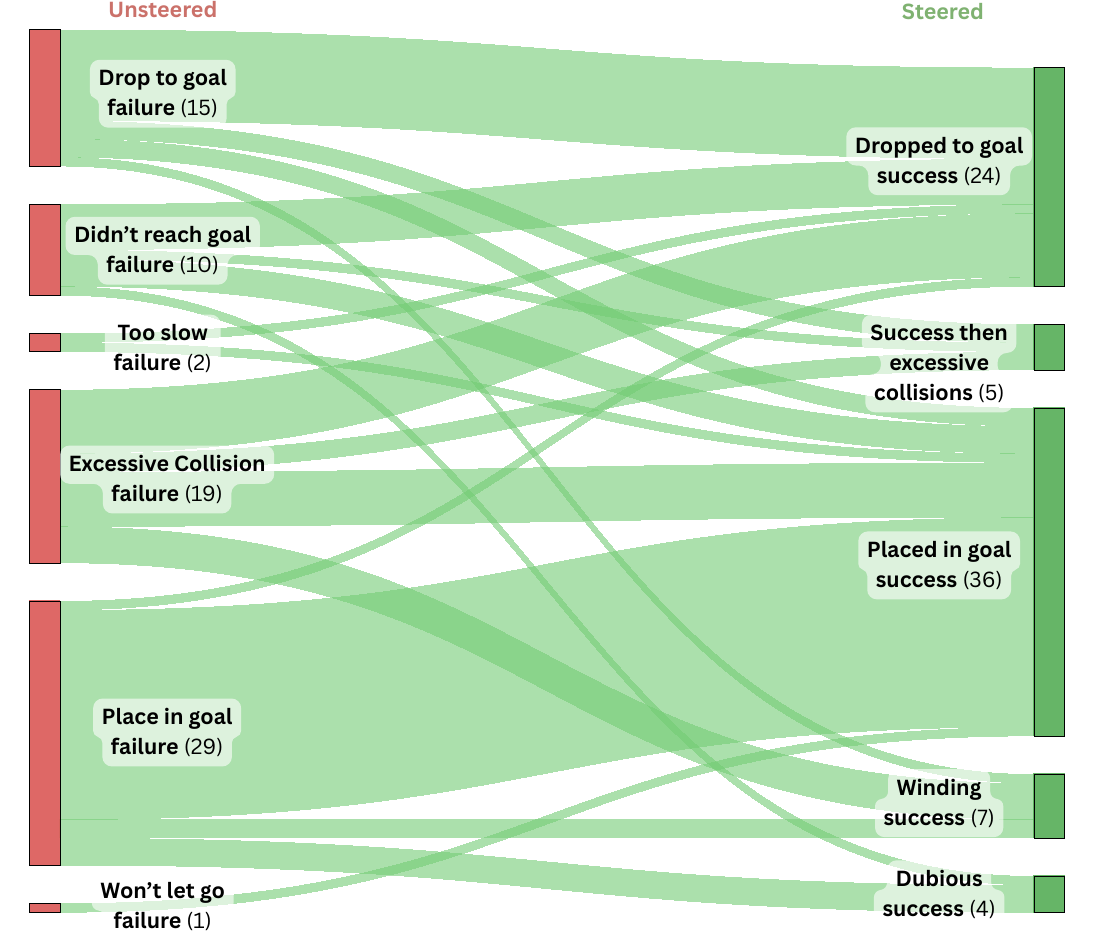}
    \caption{Sankey plot showing failure episodes switching to successful cases on \texttt{SetTable-Place}. }
    \label{fig:sankey_plot}
    \vspace{-20pt}
\end{wrapfigure}

\cref{fig:sankey_plot} showcases the exact distribution of transitions of failure types to successes through \EVE steering on MSHAB.
\texttt{EVE-Ensemble} consistently redirects catastrophic failures such as Place-in-goal failure and Excessive Collision, toward stable success modes.
This suggests that primary gains through steering arise not just from correcting rare anomalies but from restructuring the policy’s dominant error pathways. 
However, not all tasks benefit from verifier-based steering through \texttt{EVE}. 
For example, in \texttt{PrepareGroceries-Pick}, we observe minor drops in performance with the \texttt{EVE-Ensemble} configuration (see \cref{tab:mshab_results}). 
This is potentially due to the task requiring picking objects placed inside refrigerators where effective feedback from verifiers is limited.
One potential avenue to improve this is to acquire additional feedback from verifiers, such as progress estimation signals \cite{gvl}, so that the verifier can help disambiguate if the policy is stuck.  
We provide details of success and failure categories in  App. \cref{app:success_failures} and failure sankey plots for all tasks in App. \cref{app:sankey_plots}.

\section{Conclusion}
\label{sec:conclusion}

In summary, our results show that VLM-based verifier steering enhances pretrained diffusion policy performance across diverse mobile manipulation tasks.
These findings demonstrate that large VLMs can provide semantically grounded feedback to improve control in open-ended environments when combined with a guided diffusion-based action incorporator.

\bibliographystyle{splncs04}
\bibliography{main}

\newpage

\appendix
{
    \onecolumn
    \centering
    \Large
    \vspace{0.5em}\textbf{Supplementary Material} \\
    \vspace{0.5em}
}

\appendix
\renewcommand{\theHsection}{appendix.\Alph{section}}

\section{Prompts for Verifier Steering}
\label{app:prompts}

\mbox{}
\vspace{-1\baselineskip}

We provide detailed prompts that are used in the Simpler-Env task suites \cref{app:simplerenv_prompts}, ManiSkill-HAB \cref{app:mshab_prompts} and RoboTwin prompts in \cref{app:robotwin_prompts}.

\section{Verifier Ensemble Details}
\label{app:verfier_ensemble}

Recall from \cref{sec:verifier_agents} that each verifier module $V_j$ interacts with the generator policy through a verifier–specific encoding
\begin{equation}
    V_j : \Phi_j(o_t, s_t, a_{\text{gen}}) \rightarrow m_j \in \mathcal{M}_j,
    \label{eq:verifier_interface}
\end{equation}
where $\Phi_j$ defines the policy–verifier interface, $o_t$ and $s_t$ denote the current observation and proprioceptive state, $a_{\text{gen}} = \{ a_t^{(k)} \}_{k=1}^K$ is the set of candidate actions (or trajectories) from the base policy, and $m_j$ is the structured message produced by the verifier (e.g., a trajectory selection or an action primitive).

Below, we instantiate $\Phi_j$ for the Generator--Conditioned \emph{Pivot steerer} and the Generator--Agnostic \emph{Primitive steerer} used in our experiments.
We also include specific design decisions for both MSHAB and Simpler-Env environments (see App. \cref{app:task_setup} fro task details).

\textbf{Generator--Conditioned Interface: Pivot Steerer.}
The Pivot steerer is a Generator--Conditioned verifier that operates on candidate trajectories produced by the frozen base policy. In all experiments, we use $K = 40$ candidate trajectories sampled from the diffusion policy:
\begin{equation}
    a_{\text{gen}} = \big\{ a_{t:t+H}^{(k)} \big\}_{k=1}^{40},
\end{equation}
where each $a_{t:t+H}^{(k)}$ is a horizon-$H$ action sequence produced by the base policy.

The Pivot steerer interface $\Phi_{\mathrm{pivot}}$ converts $(o_t, s_t, a_{\text{gen}})$ into a compact, diverse set of trajectory visualizations in the robot’s image space, suitable for VLM prompting:
\begin{equation}
    \Phi_{\mathrm{pivot}}(o_t, s_t, a_{\text{gen}}) \;=\; \big(x, o_t, \{ \hat{\tau}^{(i)} \}_{i=1}^{K_{\text{pivot}}} \big),
\end{equation}
where $x$ is the task instruction and $\{ \hat{\tau}^{(i)} \}_{i=1}^{K_{\text{pivot}}}$ is a subset of $K_{\text{pivot}} = 5$ visually distinct trajectory overlays in the RGB image frame, derived from the original $K=40$ samples. Concretely, we proceed as follows:
\begin{enumerate}
    \item \textbf{Trajectory decoding in task space.} Each candidate sequence $a_{t:t+H}^{(k)}$ is represented either as (i) joint position deltas (for MSHAB tasks) or (ii) end-effector poses (for Simpler-Env tasks). 
    If the sequence is in joint space, we apply forward kinematics to obtain the corresponding sequence of end-effector poses $\{ T_{t+h}^{(k)} \}_{h=0}^H$. If the sequence is already given in end-effector space, we use it directly.
    \item \textbf{Projection into the RGB image frame.} For each candidate trajectory, we project the end-effector poses into the camera frame using known intrinsics and extrinsics, obtaining a 2D path in pixel coordinates. We render this path as an overlay (e.g., a polyline and/or waypoints) on top of the current RGB observation $o_t$, producing a trajectory visualization $\hat{\tau}^{(k)}$ that indicates how the end effector would move in the image plane.
    \item \textbf{Representative Trajectory Selection.} We greedily select $K_{\text{pivot}} = 5$ trajectories that are maximally diverse under cosine distance, yielding the final set $\{ \hat{\tau}^{(i)} \}_{i=1}^{K_{\text{pivot}}}$.
    \item \textbf{Prompt construction.} The interface consists of: (i) the textual task instruction $x$, (ii) the current RGB frame $o_t$, and (iii) the $K_{\text{pivot}}$ selected trajectory overlays $\{ \hat{\tau}^{(i)} \}$. These elements are serialized into a multi-modal prompt to the VLM, which is asked to select the trajectory that best completes the task.
\end{enumerate}
Given this interface, the Pivot steerer verifier $V_{\mathrm{pivot}}$ produces a message
\begin{equation}
    m_{\mathrm{pivot}} \in \mathcal{M}_{\mathrm{pivot}},
\end{equation}
which we instantiate as a discrete selection over the $K_{\text{pivot}}$ candidates (e.g., an index of the preferred trajectory) together with a natural language rationale. 
The candidate action sequence corresponding to the selected trajectory is then used to steer the base policy (lines $17-22$ of \cref{alg:eve}).

\textbf{Generator--Agnostic Interface: Primitive Steerer.}
The Primitive steerer is a Generator--Agnostic verifier and therefore does not consume generator proposals ($a_{\text{gen}} = \emptyset$ in \cref{eq:verifier_interface}). 
Instead, it directly reasons over the current observation and task instruction to select a recovery primitive from a set of predefined ones.
Its interface is given by
\begin{equation}
    \Phi_{\mathrm{prim}}(o_t, s_t, \emptyset) \;=\; \big(x, \hat{o}_t, \mathcal{A}_{\mathrm{prim}} \big),
\end{equation}
where $\mathcal{A}_{\mathrm{prim}}$ denotes the discrete set of action primitives and $\hat{o}_t$ is an augmented visual observation encoding the task goal.

Concretely, we construct $\Phi_{\mathrm{prim}}$ as follows:
\begin{enumerate}
    \item \textbf{Goal marking.} Using the task specification in MSHAB, we extract the goal location of the target object in image coordinates and overlay this location on the current RGB observation $o_t$ (e.g., by drawing a marker or highlight). The resulting goal-annotated image is denoted $\hat{o}_t$. For the Simpler-Env tasks, the task instruction is completely described only through language so we do not require goal marking.
    \item \textbf{Primitive set specification.} We define a fixed vocabulary of action primitives $\mathcal{A}_{\mathrm{prim}}$ (e.g., discrete end-effector motions and gripper commands).
    For the MSHAB tasks, we include primitives that allow for base movement and gripper action. For the Simpler-Env tasks, we include ``nudge" primitives which move the end-effector by predefined amount in specific directions.
    \item \textbf{Prompt construction.} The interface output consists of: (i) the textual task instruction $x$, (ii) the goal-marked image $\hat{o}_t$ (or just current image), and (iii) a textual description of the available primitives $\mathcal{A}_{\mathrm{prim}}$.
    These are serialized into a multi-modal prompt that asks the VLM to choose the most appropriate primitive that provides recovery.
\end{enumerate}
The Primitive steerer verifier $V_{\mathrm{prim}}$ then returns a message
\begin{equation}
    m_{\mathrm{prim}} \in \mathcal{M}_{\mathrm{prim}},
\end{equation}
which we instantiate as a single selected primitive from $\mathcal{A}_{\mathrm{prim}}$ (and optionally a natural language explanation). 
This primitive is mapped to its low-level control command and used for recovery steering using the guided diffusion incorporator in \texttt{EVE} (see \cref{sec:guided_diffusion}).

\section{Task Setup and Policy Details}
\label{app:task_setup}

In this section, we provide the task setup details for the ManiSkill-HAB task suite \cite{shukla2024maniskill} and the SimplerEnv tasks \cite{simplerenv}.

\subsection{SimplerEnv Details}
\label{app:simpler}

We use subtasks from the SimplerEnv benchmark \cite{simplerenv} which has shown strong correlation between simulator and real world evaluations.
Specifically we conduct evaluations on the following tasks across two embodiments:
\begin{itemize}
    \item \textbf{Close Drawer:} The robot is spawned in front of a cabinet which has multiple articulated drawers (top/middle/bottom).
    The robot is tasked to push and close a specific drawer in the cabinet.
    The robot can be spawned over $9$ unqiue location around the cabinet and can be tasked to close one of the $3$ drawers.
    \item \textbf{Move Object:} The robot is tasked to pick an object and place it near another specified target object.
    Each trial spawns $3$ objects in a triangular arrangement placed on the cabinet tabletop.
    \item \textbf{Place Apple in Closed Top Drawer:} The Google robot is spawned in front of a cabinet with an apple on the countertop. The robot is tasked to open the initially closed top drawer, pick up the apple and place it inside the drawer.
    \item \textbf{Carrot on Plate:} The WidowX robot is spawned in a tabletop environment containing a carrot and a plate. The robot is tasked to grasp the carrot and place it onto the plate.
    \item \textbf{Put Eggplant in Basket:} The WidowX robot is spawned in a tabletop setting and is tasked to grasp an eggplant object and accurately place it into the target basket.
    \item \textbf{Spoon on Towel:} The WidowX robot is tasked to pick up a spoon from the workspace and place it onto a towel. 
    \item \textbf{Stack Cube:} The WidowX robot is presented with two cubes on a tabletop. The robot is tasked to grasp a specific cube and stack it stably on top of another target cube.
\end{itemize}

We leverage the evaluation codebase provided in the \texttt{open-pi-zero} repository \cite{allenren_pizero}.
We use $\pi_0$ for all experiments (see App. \cref{app:pi_zero_res} for details) and use the provided checkpoints in the codebase.
For all experiments, we run $48$ rollouts spread over $3$ random seeds and ensure that the unsteered and steered runs use the exact same random seeding and episode configurations.

\subsection{ManiSkill-HAB Details}

We refer to the robot end-effector as $ee$, and its rest position as $r$.
The end-effector resting position is r = (0.5\,\text{m},\, 0\,\text{m},\, 1.25\,\text{m}) relative to the $base$.
Let $q_{\text{arm}}$ be the arm joint positions, $r_{\text{arm}}$ the arm resting
joint positions, and $\dot q_{\text{arm}}$ the arm joint velocities.
Similarly, for the torso we define $q_{\text{tor}}$, $r_{\text{tor}}$, and
$\dot q_{\text{tor}}$.
Let $v_{\text{base}}$ be the base linalear velocity in $\text{m}\,\text{s}^{-1}$
(with components $v_{\text{base},x}, v_{\text{base},y}$) and
$\omega_{\text{base}}$ the base angular velocity in
$\text{rad}\,\text{s}^{-1}$.
We initialize the robot at $(r_{\text{pos}}, r_{\text{arm}}, r_{\text{tor}})$ with
$\dot q_{\text{arm}} = 0$, $\dot q_{\text{tor}} = 0$,
$v_{\text{base}} = 0$, and $\omega_{\text{base}} = 0$,
and then add clipped Gaussian noise:
\begin{align*}
q_{\text{arm}} &\leftarrow q_{\text{arm}}
  + \mathrm{clip}(\mathcal N(0,0.1), -0.2, 0.2),\\
p_{\text{base}} &\leftarrow p_{\text{base}}
  + \mathrm{clip}(\mathcal N(0,0.1), -0.2, 0.2),\\
\theta_{\text{base}} &\leftarrow \theta_{\text{base}}
  + \mathrm{clip}(\mathcal N(0,0.25), -0.5, 0.5).
\end{align*}
The $z$-axis is “up’’ in ManiSkill3.

\textbf{Subtask definitions.}
We use the following shorthand:
\begin{align*}
d_t^a &= \lVert a_{\text{pos}} - b_{\text{pos}}\rVert_2
        \quad \text{(distance between $ee$ and its rest position)}, \\
j_k   &= \max_{1 \le i \le |q_k|} \lvert q_{k,i} - r_{k,i} \rvert
        \quad \text{(max deviation from rest for joint group $k$)}. \\
\end{align*}

We also define $C_{[0:t]}$ to be the sum of cumulative collisions from time $0$ to $t$ in N. Below, we provide the ManiSkill task definitions, success and failure criteria:
\begin{tcolorbox}[
  title={Task A: Pick[$a$, optional] $(x_{\text{pose}})$},
  colback=white,
  colframe=black!40,
  breakable
  ]
  
\textbf{Description.} Pick object $x$ from articulation $a$ (if provided).

\medskip
\textbf{Initialization.}
Spawn robot facing $x$, within $2\,$m of $x$, with noise, and without collisions.

\medskip
\textbf{Success.}
\begin{equation*}
    1_{\mathrm{grasped}(x)} \land d^{r}_{ee} \le 0.05 \land j_{\mathrm{arm}} \le 0.6
    \land 1_{\mathrm{is\_static}} \land C_{[0:t]} \le 5000
\end{equation*}

\medskip
\textbf{Failure.}
\[
    C_{[0:t]} > 5000~\text{N}.
\]
\end{tcolorbox}

\vspace{10pt}
\begin{tcolorbox}[
  title={Task B: Place[$a$, optional] $(x_{\text{pose}}, g_{\text{pos}})$},
  colback=white,
  colframe=black!40,
  breakable
]
\textbf{Description.} Place object $x$ at goal $g$ (in articulation $a$, if provided).

\medskip
\textbf{Initialization.}
Spawn with grasp pose sampled from $\mathrm{Pick}(x_{\text{pose}})$ policy, robot
facing $g$, within $2\,\mathrm{m}$ of $g$, with noise and without collisions.

\medskip
\textbf{Success.}
\begin{align*}
  \neg 1_{\text{grasped}(x)} \;\land\; d_x^g \le 0.15 \;\land\; d_{ee}^r \le 0.05 \;\land\; j_{\text{arm}} \le 0.2 \\
  \;\land\; j_{\text{tor}} \le 0.01 \;\land\; 1_{\text{is\_static}} \;\land\; C_{[0:t]} \le 7500
\end{align*}

\medskip
\textbf{Failure.}
\[
  C_{[0:t]} > 7500~\text{N}.
\]
\end{tcolorbox}

\begin{tcolorbox}[
  title={Task C: Open[$a$] $(a_{\text{pos}})$},
  colback=white,
  colframe=black!40,
  breakable
]
\textbf{Description.} Open articulation $a$ with handle at $a_{\text{pos}}$.

\medskip
\textbf{Initialization.}
Spawn the robot facing $a$. If $a$ is a fridge, sample the base pose uniformly
from the region $[0.933,-0.6] \times [1.833,0.6]$ in front of $a$; otherwise
use $[0.3,-0.6] \times [1.5,0.6]$. Add noise and ensure no collisions.

\medskip
\textbf{Success.}
Let $a_q$, $a_{q_{\max}}$, and $a_{q_{\min}}$ be the current, maximum, and minimum
joint positions for the target articulation (drawer or fridge). Define the
required opening fraction
\[
  a_{\text{ofrac}} =
  \begin{cases}
    0.75, & \text{if $a$ is a fridge},\\
    0.9,  & \text{otherwise}.
  \end{cases}
\]
We set
\[
  1_{\text{open}(a)} = 1\{\, a_q \ge a_{\text{ofrac}} (a_{q_{\max}} - a_{q_{\min}})
                           + a_{q_{\min}} \,\},
\]
and declare success if
\[
  1_{\text{open}(a)} \land d_{ee}^r \le 0.05
  \land j_{\text{arm}} \le 0.2
  \land j_{\text{tor}} \le 0.01
  \land 1_{\text{is\_static}}
  \land C_{[0:t]} \le 10\,000.
\]

\medskip
\textbf{Failure.}
\[
  C_{[0:t]} > 10\,000~\text{N}.
\]
\end{tcolorbox}

\subsubsection{Diffusion Policy Baseline}
To serve as the base policy, we train diffusion policy (DP) baselines. We use the setup from the MS-HAB paper, with a UNet backbone, a DDPM scheduler. and a 4-layer CNN for visual encoders. For visual observations, the policy relies on two onboard camera views: an egocentric head camera and a wrist-mounted (gripper) camera. For consistency, we use the same architecture and hyperparmeters for all subtasks.

\begin{table}[h]
\centering
\begin{tabular}{lc}
\hline
\textbf{Hyperparameter} & \textbf{Value} \\
\hline
Learning Rate & $0.0001$ \\
Batch Size & $256$ \\
Observation Horizon & $2$ \\
Action Horizon & $1$ \\
Prediction Horizon & $16$ \\
Diffusion Step Embedding Dim & $256$ \\
UNet Dimensions & $[256, 512, 1024]$ \\
Number of Groups & $8$ \\
Number of Training Iterations & $500,000$ \\
\hline
\end{tabular}
\caption{Diffusion Policy Hyperparameters}
\label{tab:hyperparams}
\end{table}

\textbf{Evaluation Setup.} We conduct evaluations using pretrained diffusion policy \cite{chi2025diffusion} checkpoints.
In our analysis, we consider a subset of $6$ subtasks where the base diffusion policy has a non-trivial success rate.
We report \textit{Success-Once} rates, which computes the percentage of trajectories (out of $1000$) that achieve success at least once in an episode with $200$ maximum steps.
All experiments are reported by running $24$ environments in parallel, each with $42$ episodes.
All rollouts of a particular \texttt{EVE} configuration are done in pairs of back-to-back steered and unsteered runs by ensuring the exact same random seeding.

\subsection{Robotwin-2.0 Details}
\label{app:robotwin_2.0_task_setup}
We use subtasks from the Robotwin 2.0 benchmark \cite{robotwin2}.
Specifically we conduct evaluations on the following tasks for a dual armed manipulator:
\begin{itemize}
    \item \textbf{Beat Block Hammer:} There is a hammer and a block on the table, use the arm to grab the hammer and beat the block.
    \item \textbf{Place Can Basket:} Use one arm to pick up the can, put it into the basket, and use another arm to lift the basket
    \item \textbf{Place Container Plate:} Place the container onto the plate.
    \item \textbf{Place Object Stand:} Use appropriate arm to place the object on the stand.
    \item \textbf{Move Can Pot:} There is a can and a pot on the table, use one arm to pick up the can and move it to beside the pot.
\end{itemize}

We leverage the evaluation codebase provided in the \texttt{TACO} repository \cite{taco}.
We use $\pi_{0.5}$ for all experiments and use the provided pretrained checkpoints in the codebase.
For all experiments, we run $288$ rollouts divided over $6$ random seeds and ensure that the unsteered and steered runs use the exact same random seeding and episode configurations.

\section{Baselines and Evaluation Protocol}
\label{app:baseline_evals}

We provided detailed description of baselines and evaluation procedure in this section.

\subsection{Baselines}

We provide brief description of the primary baselines we compare against on the SimplerEnv benchmark tasks (see \cref{tab:simplerenv_results}).
The baselines are defined as follows:

\begin{enumerate}
    \item \textbf{RoboMonkey} \cite{kwok2025robomonkey}: Construct a large-scale synthetic action preference dataset by applying augmentations to the BridgeV2 dataset \cite{bridgev2}. This dataset is used to finetune a LLaVA-$7B$ scale to assign higher scores to correct samples and lower scores to incorrect ones. During inference, multiple action samples are sampled from the base policy and gaussian perturbations are applied to construct the set of actions that will be scored by the verifier. The action with highest score assigned by verifier is then executed and this process is repeated at each time step.
    \item \textbf{V-GPS} \cite{nakamoto2024steering}: Trains a vlaue function using offline RL using a mix of BridgeV2 and fractal datasets \cite{o2024open}. During inference, the value function is used to generate scores (or \textit{Q-values}) for multiple actions sampled by the base policy. This is followed by an action re-ranking step where the final action is sampled from a “re-ranked” categorical distribution obtained by computing a temperature-controlled softmax over Q-values.
\end{enumerate}

\subsection{Evaluation Details}

\textbf{SimplerEnv Evals}. For \EVE, we use ensemble weights (\textit{Pivot:Primitive}) by performing a search for \textit{Pivot} weights with values from: $[0.1,0.2,0.3,0.5]$ for each task.
For RoboMonkey, we sample $9$ action candidates from the base policy and augment to $32$ samples using gaussian perturbations.
For VGPS, we sample $10$ actions from the base policy and use a softmax temperature of $0.1$ for action re-ranking. 
We additionally provide detailed hyperparameters in \cref{tab:hyperparams_simpler}.
These hyperparameters are kept constant across tasks.

\textbf{MSHAB parameters.} We outline hyperparameters used in Maniskill-HAB evaluations in \cref{tab:hyperparams_mshab}.

\textbf{Robotwin parameters.} We outline hyperparameters used in Robotwin 2.0 evaluations (see \cref{app:robotwin_result}) in \cref{tab:hyperparams_simpler}.

\section{Extended Results and Diffusion Guidance}
\label{app:extended_results_and_evals}

\subsection{RoboTwin 2.0 Task Suite Results}
\label{app:robotwin_result}

\textbf{Evaluation Details.} For all experiments presented in \cref{sec:long_horizon_tasks} of main text, we use the \texttt{EVE-Ensemble} variant.
Each task was run over $6$ seeds with $48$ rollouts per seed.
All evaluated tasks use the \textit{exact same hyper-parameters} listed in \cref{tab:hyperparams_simpler}.
These were selected by searching over guidance scale: $[0.5-5]$ (step=$0.5$), MMD threshold: $[0.8-1.2]$ (step=$0.1$), Pivot weight: $[0.1-0.9]$ (step=$0.2$).

\subsection{\texttt{EVE} Steering with Flow-Based VLAs}
\label{app:pi_zero_res}

In the main paper, we design an action incorporator module that leverages guided diffusion (see \cref{sec:guided_diffusion}) to steer diffusion policies, but we can also apply \texttt{EVE} to base policies trained with conditional flow matching \cite{lipman2022flow}.
Specifically we present an adaptation of \texttt{EVE} to a large flow-based VLA policy $\pi_0$ \cite{pizero} in the following section and test performance on the SimplerEnv benchmark (see \cref{tab:simplerenv_results}). 
Additionally we also apply the same technique to the more recent $\mathbf{\pi_{0.5}}$ policy and present results on Robotwin 2.0 in \cref{app:robotwin_result}.

\textbf{Flow-matching policies.}
Given observation $o_t$, a flow policy generates an action chunk by first sampling Gaussian noise
$A_t^{0} \sim \mathcal N(0, I)$ and then integrating the learned velocity field $v_\pi$ over a
``flow time'' variable $\tau \in [0,1]$:
\begin{equation}
A_t^{\tau+\Delta} = A_t^{\tau} + \Delta\, v_\pi(A_t^{\tau}, o_t, \tau), 
\qquad \Delta = \tfrac{1}{n},
\label{eq:flow-policy}
\end{equation}
where $n$ is the number of denoising steps.
The final action is $A_t^{1}$ after $n$ Euler steps of Eq.~\eqref{eq:flow-policy}.

\textbf{Guided Inference for Flow Policies.}
To steer the flow generation towards a specified action sequence $A_{\text{ref}}$ (e.g., an action suggested by the \texttt{EVE} verifier system), we use the guided inference scheme proposed in \cite{black2025real}. 
Let $v_{\pi}(A_\tau, o, \tau)$ denote the velocity predicted by the policy at flow step $\tau$.
We first compute the estimated terminal (clean) action $\hat{A}_1$:
\begin{equation}
    \hat{A}_1 = A_\tau + (1 - \tau) v_{\pi}(A_\tau, o, \tau).
    \label{eq:estimated_clean}
\end{equation}
We define the guidance objective as minimizing the squared error between this estimate and the reference action $A_{\text{ref}}$.
We can compute the gradient of the loss $\mathcal{L} = \frac{1}{2}||\hat{A}_1 - A_{\text{ref}}||^2$ (similar to \cref{eq:alignment_objective_action}):
\begin{equation}
    \nabla_{v} \mathcal{L} = (1 - \tau)(\hat{A}_1 - A_{\text{ref}}).
    \label{eq:l2_flow_graident}
\end{equation}
We then adjust the velocity using a guidance scale $\gamma$ and perform the standard Euler integration step:
\begin{align}
    \hat{v}_{\tau} &= v_{\pi}(A_\tau, o, \tau) - \gamma \nabla_{v} \mathcal{L}, \\
    A_{\tau+\Delta \tau} &= A_\tau + \Delta \tau \, \hat{v}_{\tau}.
    \label{eq:guided_flow}
\end{align}

\textbf{Integrating \texttt{EVE} with Flow Policies.}
When the base policy is a diffusion model, we use \cref{alg:eve} which performs guided diffusion using the verifier-ensemble action output in lines $15-23$ via a DDPM-style reverse process.
For a flow-based base policy, the outer structure of \cref{alg:eve} is unchanged:
we still draw $K$ candidate action chunks from the frozen base policy (lines $1-4$), run intervention detection using
MMD-based detector (lines $5-7$), and run the verifier ensemble to obtain an aggregated
correction signal $\tilde{m}$ (lines $8-13$).
The only modification is that lines $15-23$ are replaced by the guided flow integration of
Eq.~\eqref{eq:guided_flow}, where $\hat{v}_{\tau}$ is computed by setting $A_{\text{ref}}$ to $\tilde{m}$ in Eq.~\eqref{eq:l2_flow_graident}.
Thus EVE can be applied to both diffusion and flow-matching policies with a minimal change to the
inference procedure.
For all experiments, we use the exact same \textit{Primitive} and \textit{Pivot} steerers as defined in \cref{sec:implementation_details}.

\section{Detailed Ablations}
\label{sec:detailed_ablations}
\begin{figure*}[htbp]
  \centering
  \begin{subfigure}[t]{0.30\textwidth}
      \includegraphics[width=\linewidth]{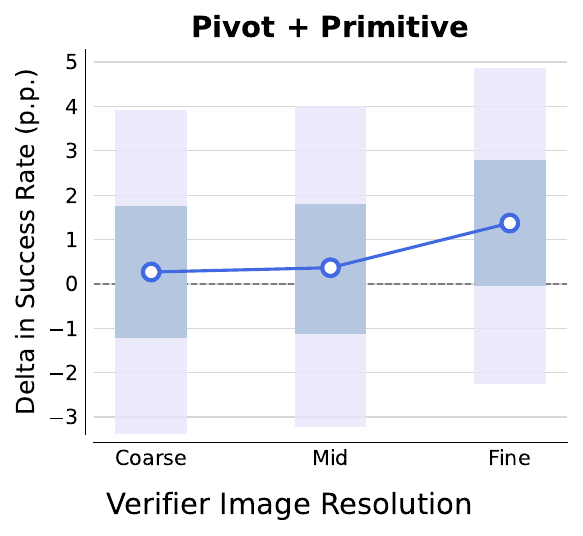}
      \caption{}
      \label{fig:img_res_ablate}
  \end{subfigure}
  \hfill
  \begin{subfigure}[t]{0.30\textwidth}
      \includegraphics[width=\linewidth]{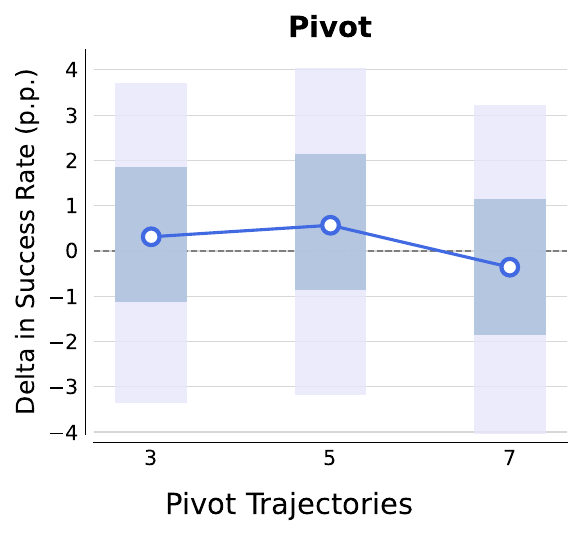}
      \caption{}
      \label{fig:pivot_traj_ablation}
  \end{subfigure}
  \hfill
  \begin{subfigure}[t]{0.30\textwidth}
      \includegraphics[width=\linewidth]{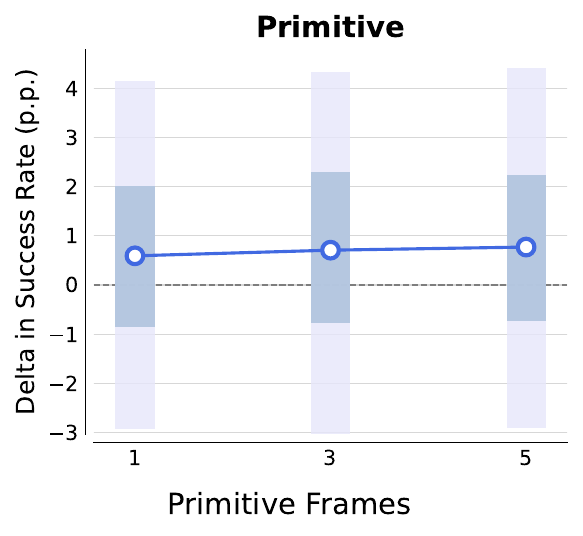}
      \caption{}
      \label{fig:primitive_frames_ablation}
  \end{subfigure}

  \vspace{-2mm}
  \caption{\EVE ablations on the MSHAB task suite (verifier configurations).}
  \label{fig:eve_ablation_verifier}
\end{figure*}
\textbf{Verifier Image Resolution.}
In this experiment, we re-render images at a higher resolution of $256$p and $512$p from the simulator and pass them to the verifier ensemble. 
From \cref{fig:img_res_ablate}, we see that using images with higher visual resolution enhances performance significantly.
We note that higher resolution images are important for contact-rich mobile manipulation tasks in MSHAB and enable the verifier to provide fine-grained action feedback.

\textbf{Verifier Information Ablations.}
In this experiment, we analyze the density of information that is passed to the individual verifiers through their respective policy-verifier interfaces (see $\Phi_j$ defined in \cref{sec:methodology}).
For the \textit{Pivot} verifier, we ablate the number of trajectories that the verifier can select from.
From \cref{fig:pivot_traj_ablation} we observe an increase as the number of drawn trajectories is increased from $3$ to $5$.
But increasing the number of trajectories to $7$ leads to degradation in the performance, potentially because the VLM can no longer effectively discern between the trajectories.
For the \textit{Primitive} verifier, we ablate the number of history frames that are passed to the VLM.
In \cref{fig:primitive_frames_ablation}, we ablate the number of history frames that are passed to the \textit{Primitive} verifier for recovery primitive selection.
From the results, we observe that increasing the frame history doesn't affect performance significantly.
We hypothesize this is because the robot doesn't have very large movements in adjacent frames once it approaches the receptacle from which it needs to either pick or place an object.

\section{Intervention Detection using Mean Maximum Discrepancy Scores}
\label{app:mmd_detail}

\begin{figure*}[htbp]
  \centering
  
  \begin{subfigure}[t]{0.32\textwidth}
    \includegraphics[width=\linewidth]{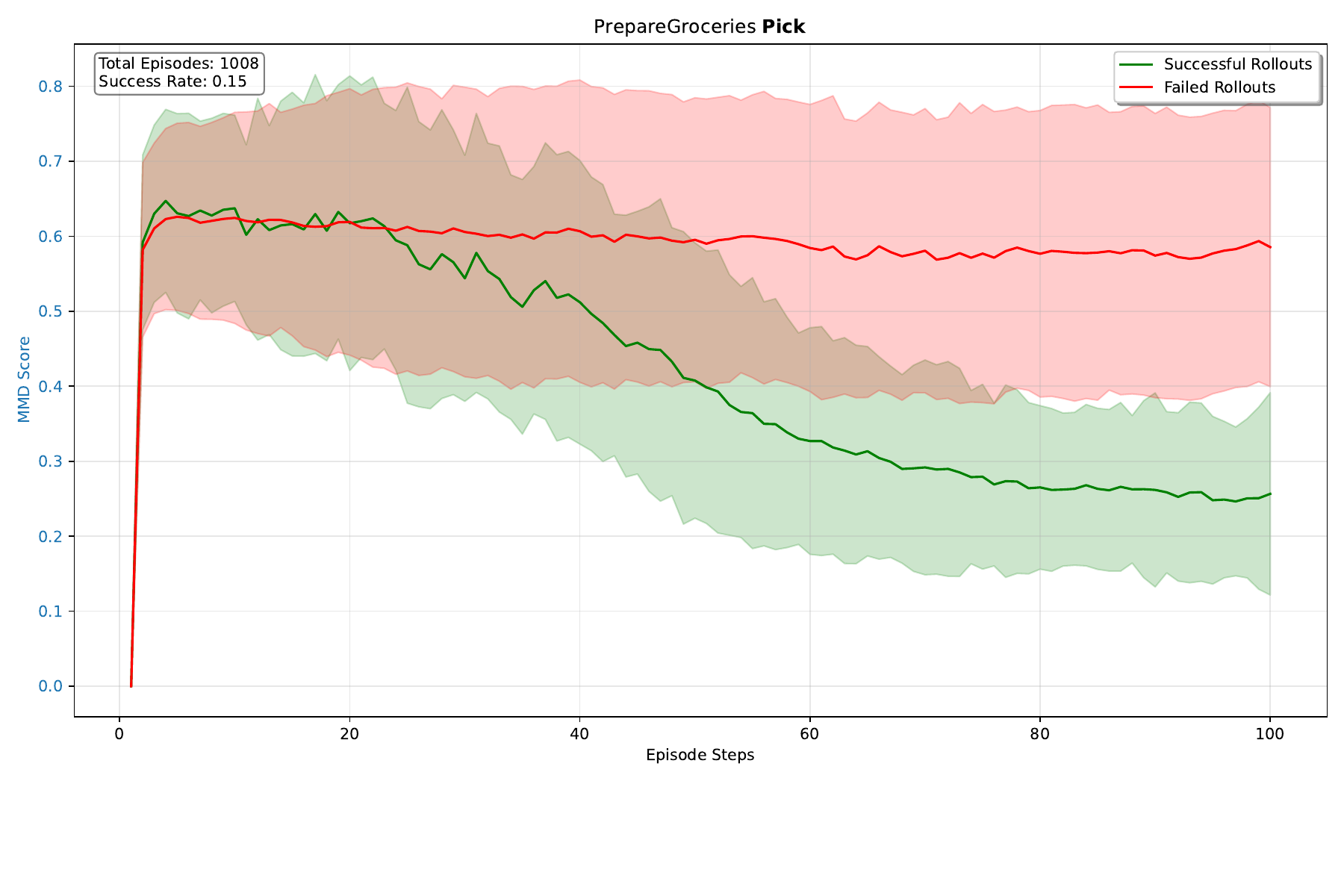}
    \caption{}
    \label{fig:prepare_pick_mmd}
  \end{subfigure}
  \hfill
  \begin{subfigure}[t]{0.32\textwidth}
    \includegraphics[width=\linewidth]{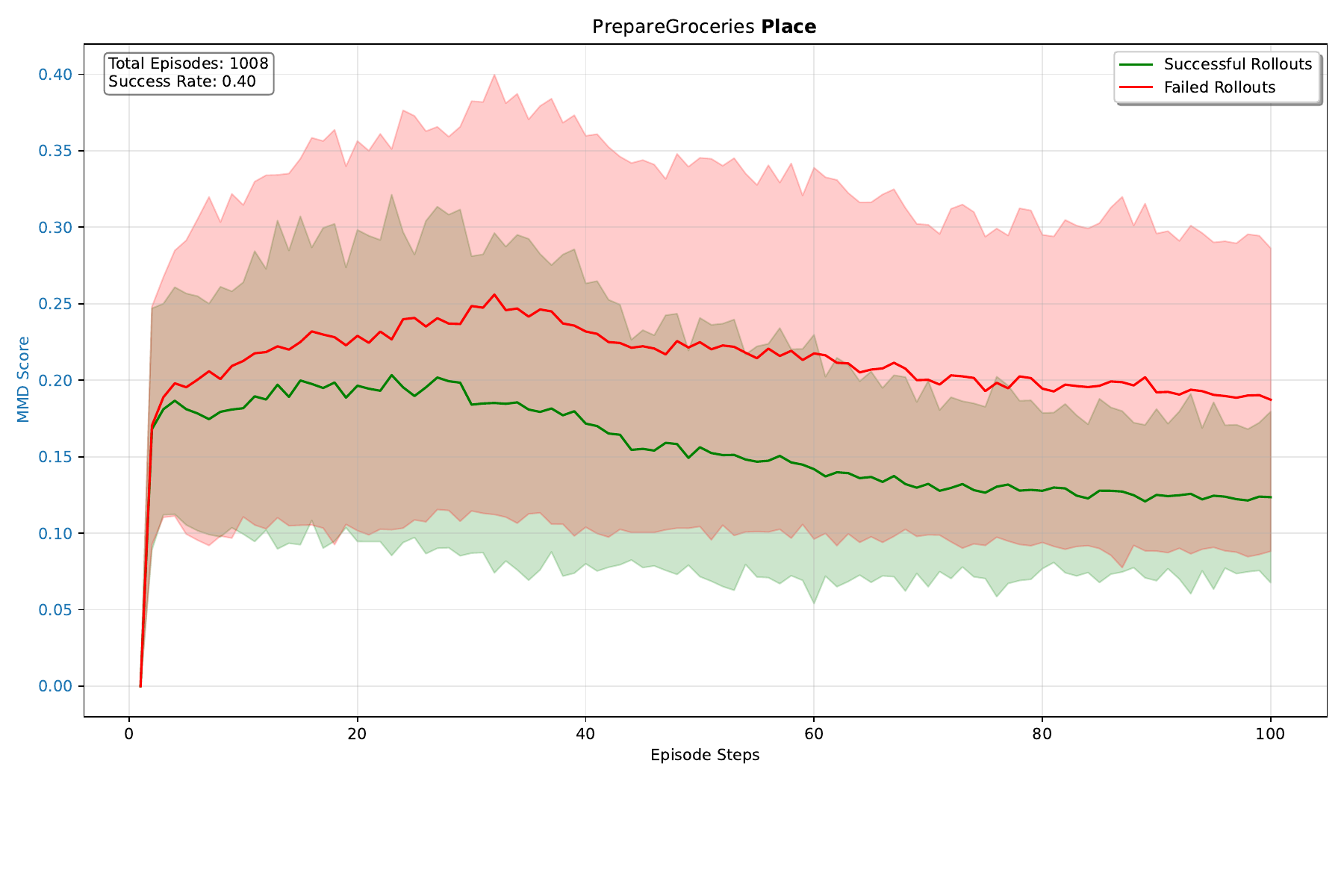}
    \caption{}
    \label{fig:prepare_place_mmd}
  \end{subfigure}
  \hfill
  \begin{subfigure}[t]{0.32\textwidth}
    \includegraphics[width=\linewidth]{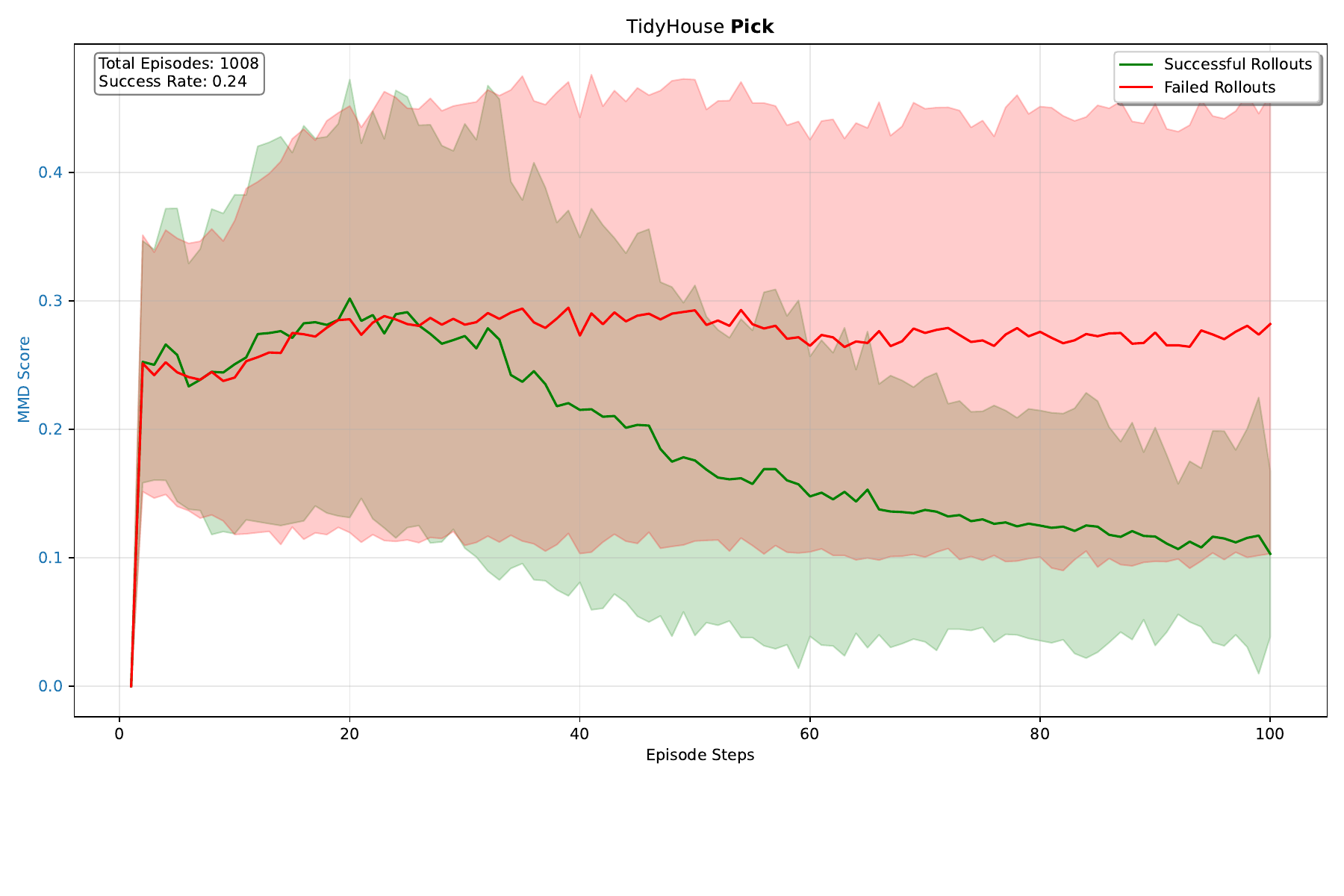}
    \caption{}
    \label{fig:tidy_pick_mmd}
  \end{subfigure}
  \vspace{-2mm}
  \begin{subfigure}[t]{0.32\textwidth}
    \includegraphics[width=\linewidth]{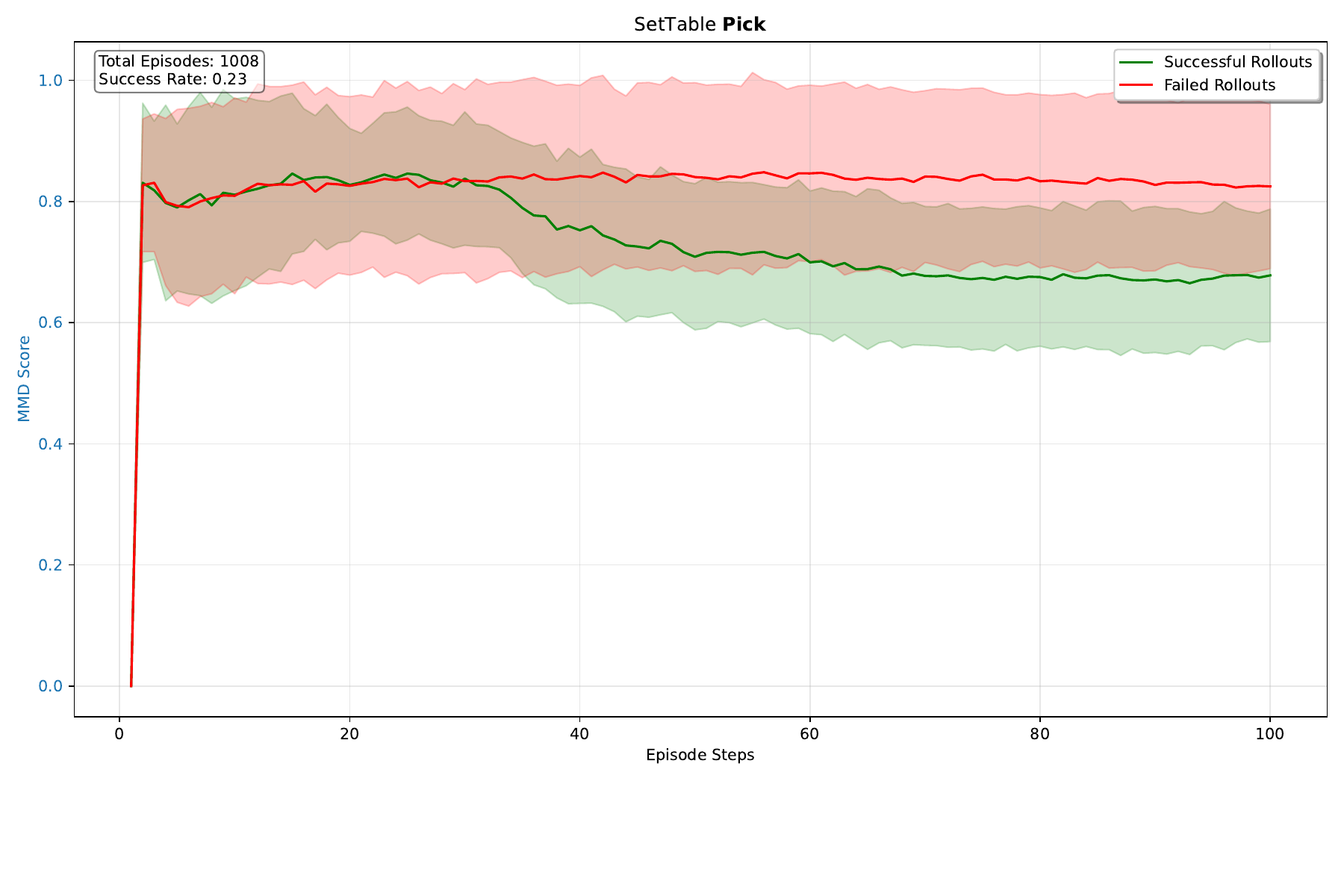}
    \caption{}
    \label{fig:table_pick_mmd}
  \end{subfigure}
  \hfill
  \begin{subfigure}[t]{0.32\textwidth}
    \includegraphics[width=\linewidth]{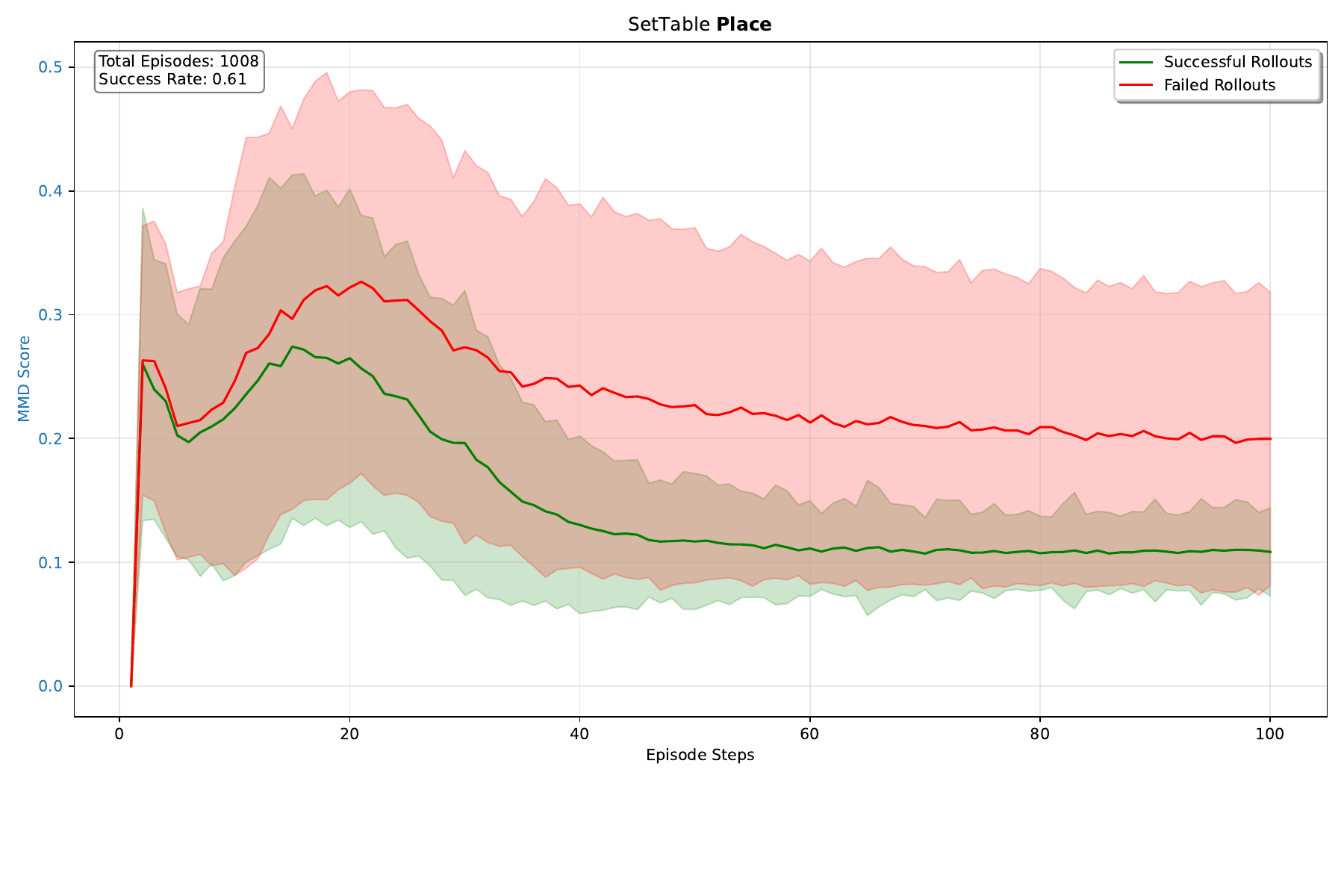}
    \caption{}
    \label{fig:table_place_mmd}
  \end{subfigure}
  \hfill
  \begin{subfigure}[t]{0.32\textwidth}
    \includegraphics[width=\linewidth]{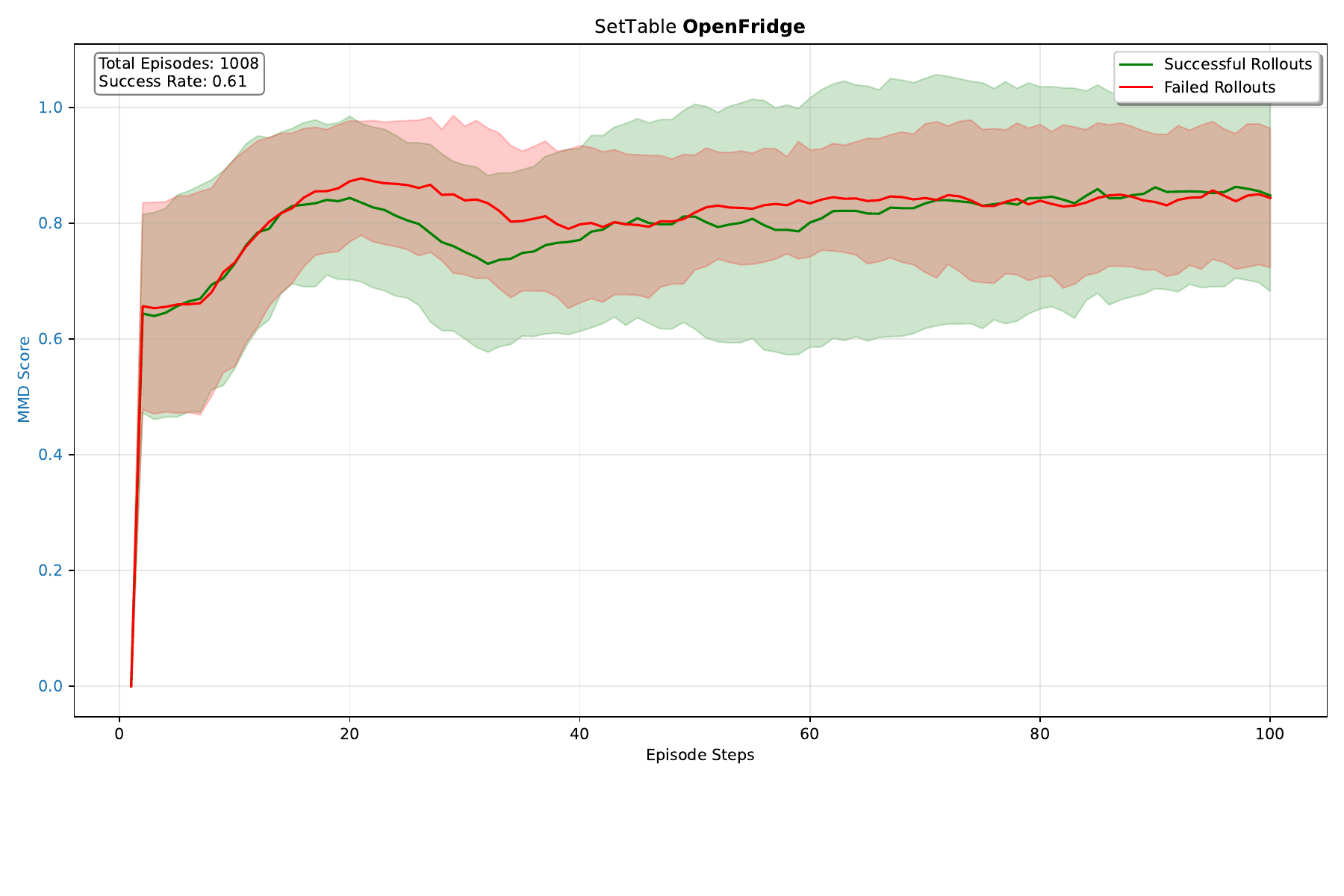}
    \caption{}
    \label{fig:table_open_mmd}
  \end{subfigure}
  \caption{Base policy MMD scores for successful and failed rollouts on $6$ mobile manipulation tasks from the MS-HAB \cite{shukla2024maniskill} benchmark. Across tasks, we consistently find that MMD scores discriminate between failed and successful rollouts.}
  \label{fig:mmd_distributions}
\end{figure*}

To detect distribution shifts and potential erratic behaviors during test-time deployment, we employ the Maximum Mean Discrepancy (MMD) metric. 
This formulation is based on the STAC metric proposed by Agia et al.~\cite{agia2024unpacking}, which quantifies the distance between the distribution of action trajectories generated at the current timestep $t$ and those generated at the future timestep $t+k$. 
In \cref{fig:mmd_distributions} we qualitatively show that MMD scores consistently discern between success and failure rollouts.
This section details the mathematical formulation of the temporal consistency check used in \cref{sec:intervention_detection}.

\noindent \textbf{Marginal Action Distributions:} We consider a receding horizon control setting where the policy $\pi_\theta$ predicts a sequence of actions (an action chunk) of length $H$. Let $k$ denote the execution horizon (the number of steps the robot executes before replanning).
At timestep $t$, the policy generates a distribution of trajectories. Following the formulation in~\cite{agia2024unpacking}, we isolate the segment of this trajectory that overlaps with the next planning step $t+k$. The overlapping temporal window has a length of $H-k$. 

We define the two marginal distributions over this overlapping window as follows:

\begin{enumerate}
    \item $\overline{\pi}_t$: The distribution of action sequences generated at time $t$, restricted to the window $[t+k, t+H-1]$. Formally, $\overline{\pi}_t := p(a_{t+k:t+H-1} \mid o_t, s_t)$.
    \item $\tilde{\pi}_{t+k}$: The distribution of action sequences generated at time $t+k$, restricted to the same window $[t+k, t+H-1]$. Formally, $\tilde{\pi}_{t+k} := p(a_{t+k:t+H-1} \mid o_{t+k}, s_{t+k})$.
\end{enumerate}

\hspace{-10pt} Under nominal conditions, the policy's plan at time $t$ for the future window should remain consistent with the updated plan generated at $t+k$. 
A high divergence between $\overline{\pi}_t$ and $\tilde{\pi}_{t+k}$ indicates erratic behavior or distribution shift.

\noindent \textbf{Maximum Mean Discrepancy (MMD):}
We measure the distance between these two distributions using the squared MMD in a Reproducing Kernel Hilbert Space (RKHS) $\mathcal{H}$ associated with a kernel function $k(\cdot, \cdot)$. The squared population MMD is defined as~\cite{agia2024unpacking}:

\begin{equation}
\begin{aligned}
    D^2(\overline{\pi}_t, \tilde{\pi}_{t+k}) &= \mathbb{E}_{\mathbf{x}, \mathbf{x}' \sim \overline{\pi}_t} [k(\mathbf{x}, \mathbf{x}')] \\
    &+ \mathbb{E}_{\mathbf{y}, \mathbf{y}' \sim \tilde{\pi}_{t+k}} [k(\mathbf{y}, \mathbf{y}')] \\
    &- 2\mathbb{E}_{\mathbf{x} \sim \overline{\pi}_t, \mathbf{y} \sim \tilde{\pi}_{t+k}} [k(\mathbf{x}, \mathbf{y})]
\end{aligned}
\label{eq:mmd_population}
\end{equation}
where $\mathbf{x}, \mathbf{y}$ represent the flattened vectors of the action sequences $a_{t+k:t+H-1}$ sampled from their respective distributions.
We employ the Radial Basis Function (RBF) as the kernel function.

\noindent\textbf{Empirical Estimation:}
Since the analytical densities of the diffusion policy are intractable, we approximate Equation \ref{eq:mmd_population} using a finite batch of samples, consistent with the STAC implementation~\cite{agia2024unpacking}. We draw $B$ samples from the policy at timestep $t$ (denoted as $X = \{\mathbf{x}_i\}_{i=1}^B$) and $B$ samples at timestep $t+k$ (denoted as $Y = \{\mathbf{y}_j\}_{j=1}^B$).
The empirical MMD estimate $\hat{D}$ is computed as:

\begin{equation}
    \hat{D}(\overline{\pi}_t, \tilde{\pi}_{t+k}) = \frac{1}{B^2} \sum_{i=1}^B \sum_{j=1}^B k(\mathbf{x}_i, \mathbf{x}_j) + \frac{1}{B^2} \sum_{i=1}^B \sum_{j=1}^B k(\mathbf{y}_i, \mathbf{y}_j) - \frac{2}{B^2} \sum_{i=1}^B \sum_{j=1}^B k(\mathbf{x}_i, \mathbf{y}_j)
    \label{eq:mmd_equation}
\end{equation}
This empirical estimate provides a differentiable and computationally efficient signal of temporal inconsistency. During inference, if the MMD score computed across adjacent timesteps exceeds the threshold $\tau$, we trigger the intervention mechanism described in \cref{sec:intervention_detection}.

\section{vLLM Inference Details}
\label{app:vllm_details}

We run all VLM-based verifier inferencing using the vLLM library \cite{kwon2023efficient}.
vLLM provides highly optimized KV cache management using paged attention which enables high throughput handling of concurrent API requests from multiple clients.
Specifically, we run each \texttt{Qwen-2.5-VL-72B} instance on a single node of $8$ NVIDIA $48$GB A$40$ GPUs.
Each vLLM single node server is run with a tensor-parallel rank of $8$.
We limit gpu memory utilization on each server to $0.85$ to prevent crashes due to large burst in number of verification API requests from parallel simulator environments. 

\section{Task Success and Failure Categories}
\label{app:success_failures}

We use the trajectory categorization system proposes in MSHAB \cite{shukla2024maniskill}. 
The authors define various event-based heuristics which enables automated categorization of policy trajectory rollouts into specific success and failure modes based on the chronological sequence of events.
In \cref{tab:place_failure_cats,tab:open_failure_cats} we provide the qualitative descriptions for the modes associated with the Place and Open subtasks (reproduced from the original MSHAB paper).

\section{Failure Sankey Plots}
\label{app:sankey_plots}

For all success and failure definitions please refer to \cref{app:success_failures} for details.
In this section, we provide additional sankey plots for additional analysis on \EVE verifier steering.

\begin{figure}[t]
  \centering
  \resizebox{0.8\linewidth}{!}{%
  \begin{minipage}{\linewidth}
    \centering

    \begin{subfigure}{0.48\linewidth}
      \centering
      \includegraphics[width=\linewidth]{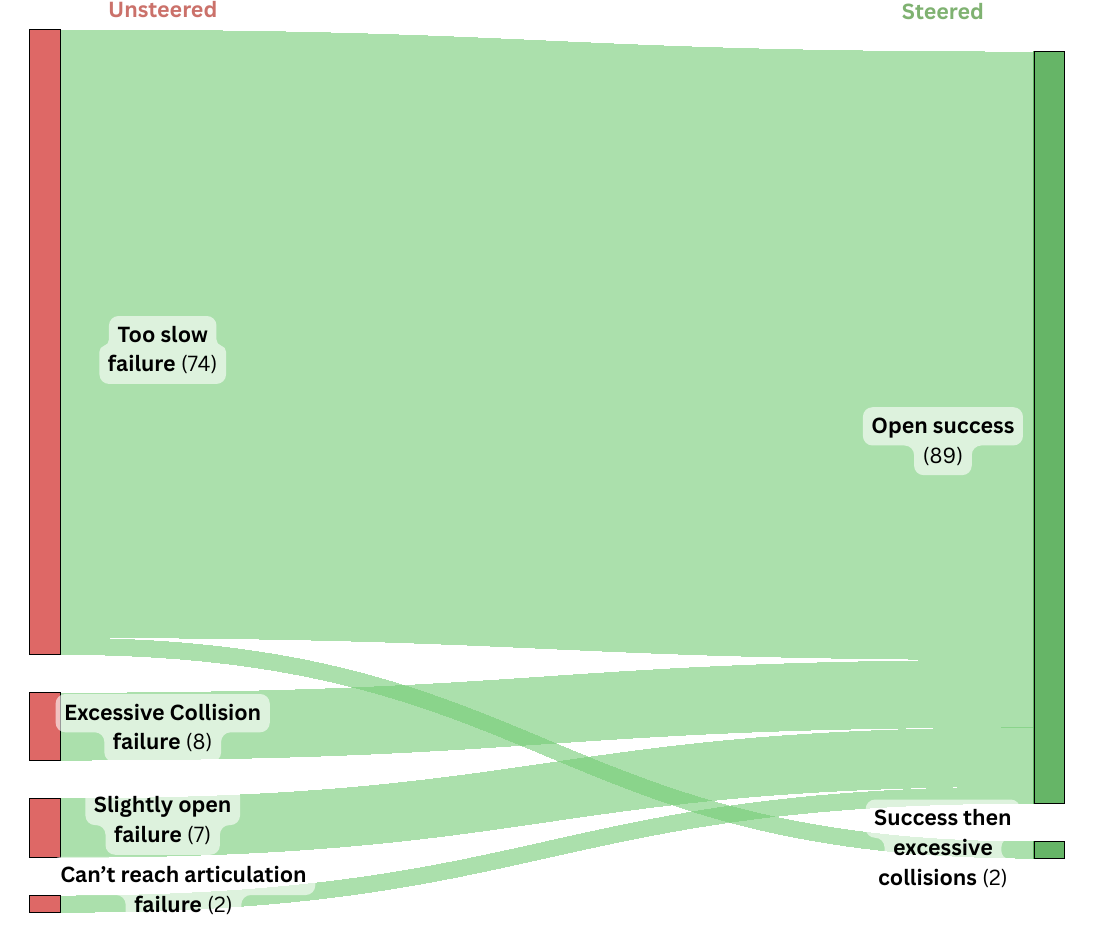}
      \caption{Failure episodes switching to successful cases on \texttt{SetTable-OpenFridge}.}
      \label{fig:set_table_open_fridge_sankey_pivot_primitive}
    \end{subfigure}
    \hfill
    \begin{subfigure}{0.48\linewidth}
      \centering
      \includegraphics[width=\linewidth]{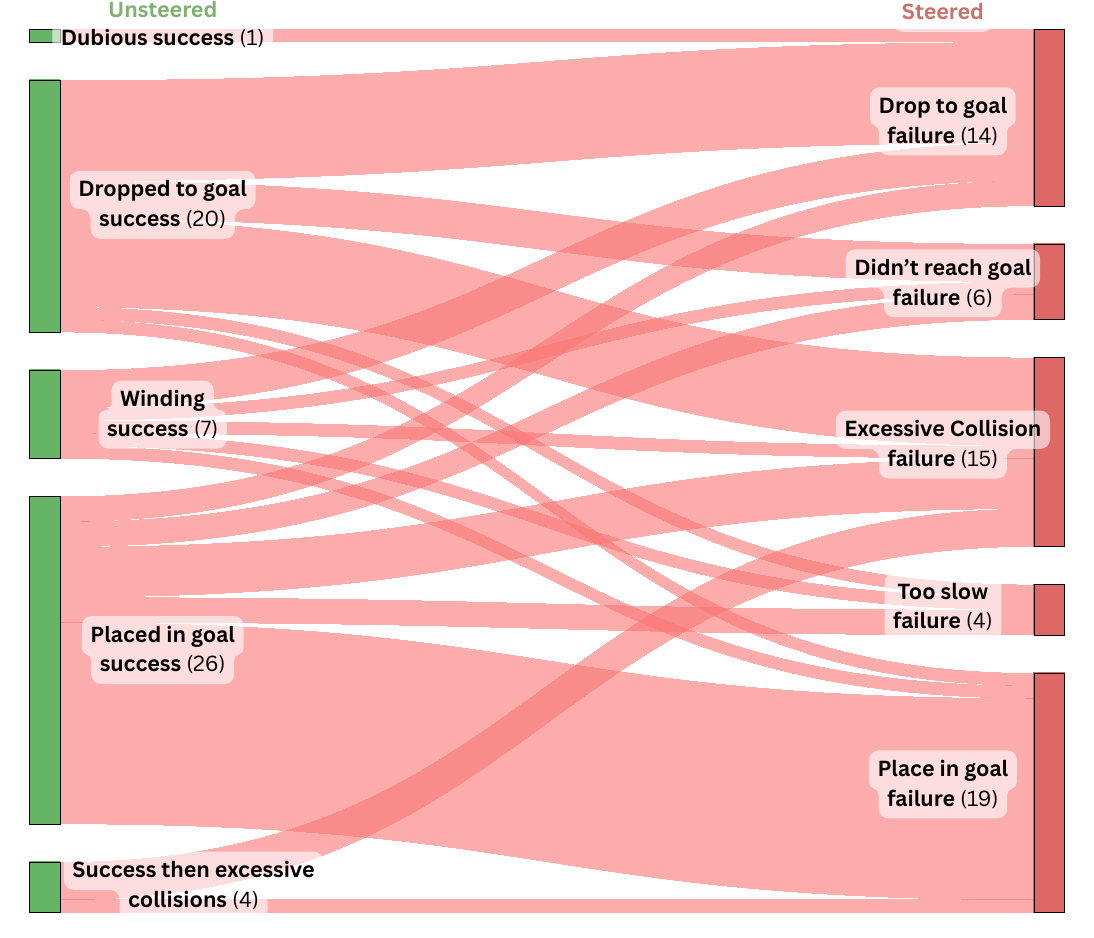}
      \caption{Success to failure episode switches on \texttt{SetTable-Place}. See \cref{fig:sankey_plot} for inverted analysis.}
      \label{fig:set_table_place_sf_sankey_pivot_primitive}
    \end{subfigure}

  \end{minipage}
  }

  \caption{Sankey visualizations of verifier steering effects on \texttt{SetTable} tasks using \texttt{EVE-Ensemble} verifier steering.}
  \label{fig:set_table_sankey_combined}
\end{figure}

\textbf{Failure-to-Success Transitions.}
In \cref{fig:set_table_open_fridge_sankey_pivot_primitive} we observe that for articulation-focused tasks such as \texttt{SetTable-OpenFridge}, steering nearly eliminates ``too slow’’ and ``cant reach" articulation failures, with almost all trajectories ending as open successes.
Overall, steering behaves as a robust correction mechanism that enables recovery from catastrophic failure scenarios.

\textbf{Success-to-Failure Transitions.} \cref{fig:set_table_place_sf_sankey_pivot_primitive} analyzes how verifier-based steering perturbs successful \texttt{SetTable-Place} rollouts by re-running unsteered success episodes with steering enabled and categorizing the resulting failures. 
We observe that high-quality ``placed in goal” successes are reclassified as ``place in goal failure” or ``excessive collision failures" which indicates that slight errors in execution can lead to slowly decaying failures (see \cref{tab:place_failure_cats}), even if the object is correctly placed at the target location initially.
This suggests that the verifier systems need to be improved to provide precise feedback which can help prevent such delayed failures in the policy rollout.

\section{Verifier Inference Overhead}
\label{app:verifier_overhead}

\begin{table}[h]
    \centering
    \begin{tabular}{c|c|c}
    \toprule
    \textbf{Environments} & \textbf{Primitive Steering} & \textbf{Ensemble Throughput} \\
    \midrule
    1  & 0.11 & 0.04 \\
    10 & 0.26 & 0.10 \\
    20 & 0.34 & 0.10 \\
    40 & 0.41 & 0.12 \\
    \bottomrule
    \end{tabular}
    \caption{Average Throughput (server responses/second) comparison across environments for Primitive Steering and Ensemble methods on a single vllm server.}
    \label{tab:steering_latency}
\end{table}

In this section, we analyze verifier inference overheads in terms of inference latency and system throughput across different configurations and scale in the MSHAB setting. The goal is to demonstrate how the \texttt{EVE} system latency scales with respect to server resources to demonstrate that compute can effectively be shared between parallel policy rollouts simultaneously running \texttt{EVE} steered inference. 

\textbf{vLLM Request Batching}
Table~\ref{tab:steering_latency} effectively demonstrates that as the number of environments increases, the average latency (average time a robot spends waiting for a vllm server response) for an environment increases, yet, as multiple requests are running simultaneously, throughput increases ($0.04$ to $0.12$ responses per second) by $\mathbf{3\times}$.
We observe that increasing the batch size (number of parallel environments) effectively improves system throughput.
For the Primitive steerer, increasing the environment count from 1 to 40 results in a near $\mathbf{4\times}$ increase in throughput ($0.11$ to $0.41$ responses/s). This demonstrates that the underlying vLLM serving infrastructure effectively leverages batching to amortize the cost of large model inference.

\begin{wraptable}{r}{0.38\textwidth}
    \vspace{-8pt}
    \centering
    \begin{tabular}{c|c}
    \toprule
    \textbf{Servers} & \textbf{Eval.\ Time} \\
     & \textbf{(min)} \\
    \midrule
    1 & 91.68 \\
    2 & 62.88 \\
    4 & 37.20 \\
    6 & 30.35 \\
    \bottomrule
    \end{tabular}
    \caption{Experiment time across different server counts for $1008$ rollouts in MSHAB.}
    \label{tab:server_performance}
    \vspace{-10pt}
\end{wraptable}

\textbf{System Scalability}
To mitigate the inference bottleneck during large-scale evaluation, we experiment with using multiple inference servers.
For this experiment we fix the total number of episodes to $1008$ across $24$ envs ($42$ episodes each) as in the main evaluations.
We can now distribute the $24$ envs to multiple servers.
Table~\ref{tab:server_performance} illustrates the reduction in total evaluation time as we scale the compute resources from 1 to 6 servers.
We observe a strong linear scaling trend. 
By increasing the server count from 1 to 6, the evaluation time for a fixed set of rollouts drops from $\sim91$ minutes to $\sim30$ minutes.

\section{Qualitative Examples}
\label{app:qual_examples}

\begin{figure}[htbp]
    \centering
    \begin{subfigure}[b]{0.7\textwidth}
        \centering
        \includegraphics[width=\textwidth]{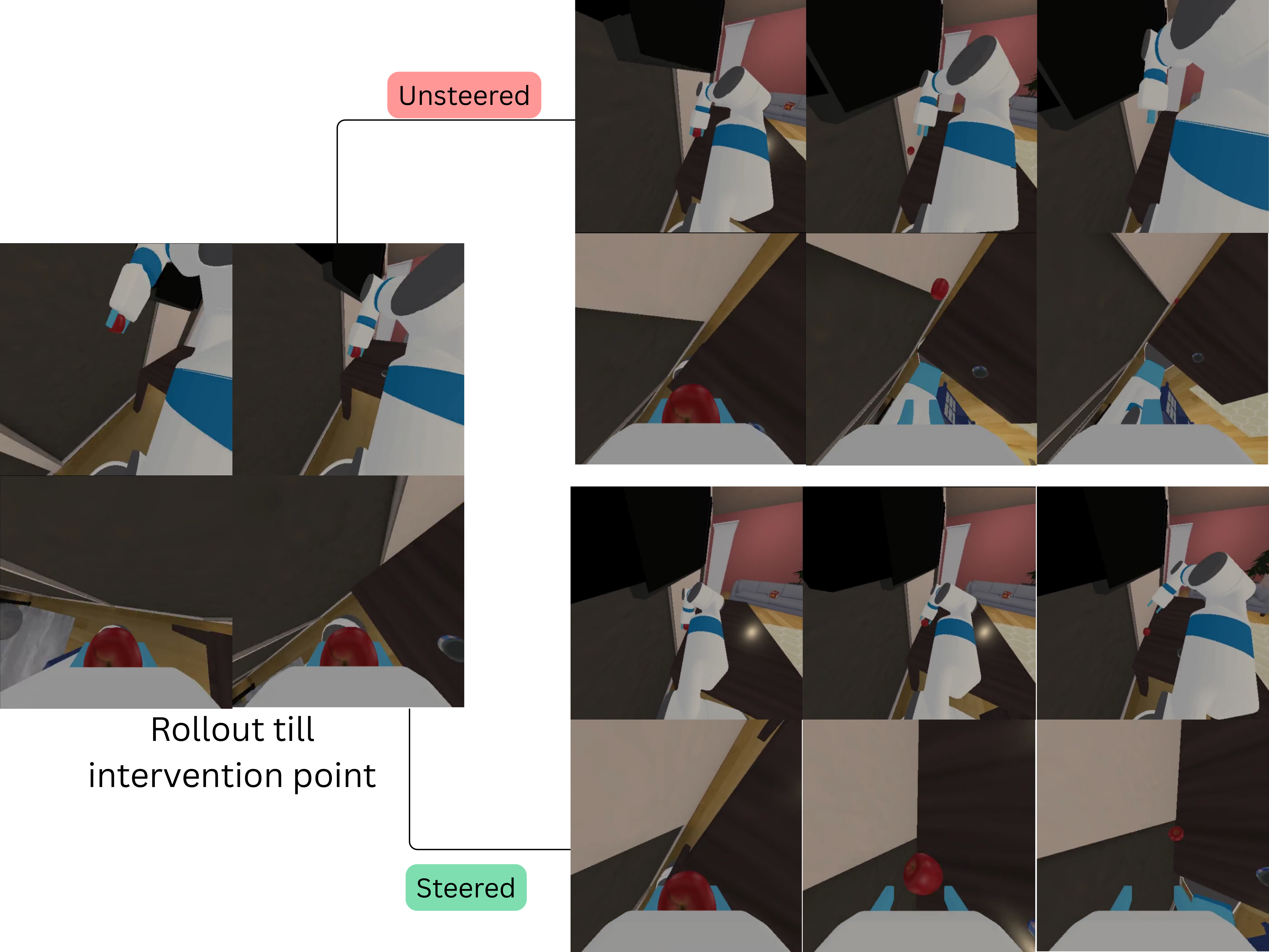}
        \caption{Visual comparison of rollouts. The left side shows the state upto the intervention point, leading to Unsteered (top) and Steered (bottom) outcomes.}
        \label{fig:intervention}
    \end{subfigure}
    \begin{subfigure}[b]{0.7\textwidth}
        \centering
        \includegraphics[width=\textwidth]{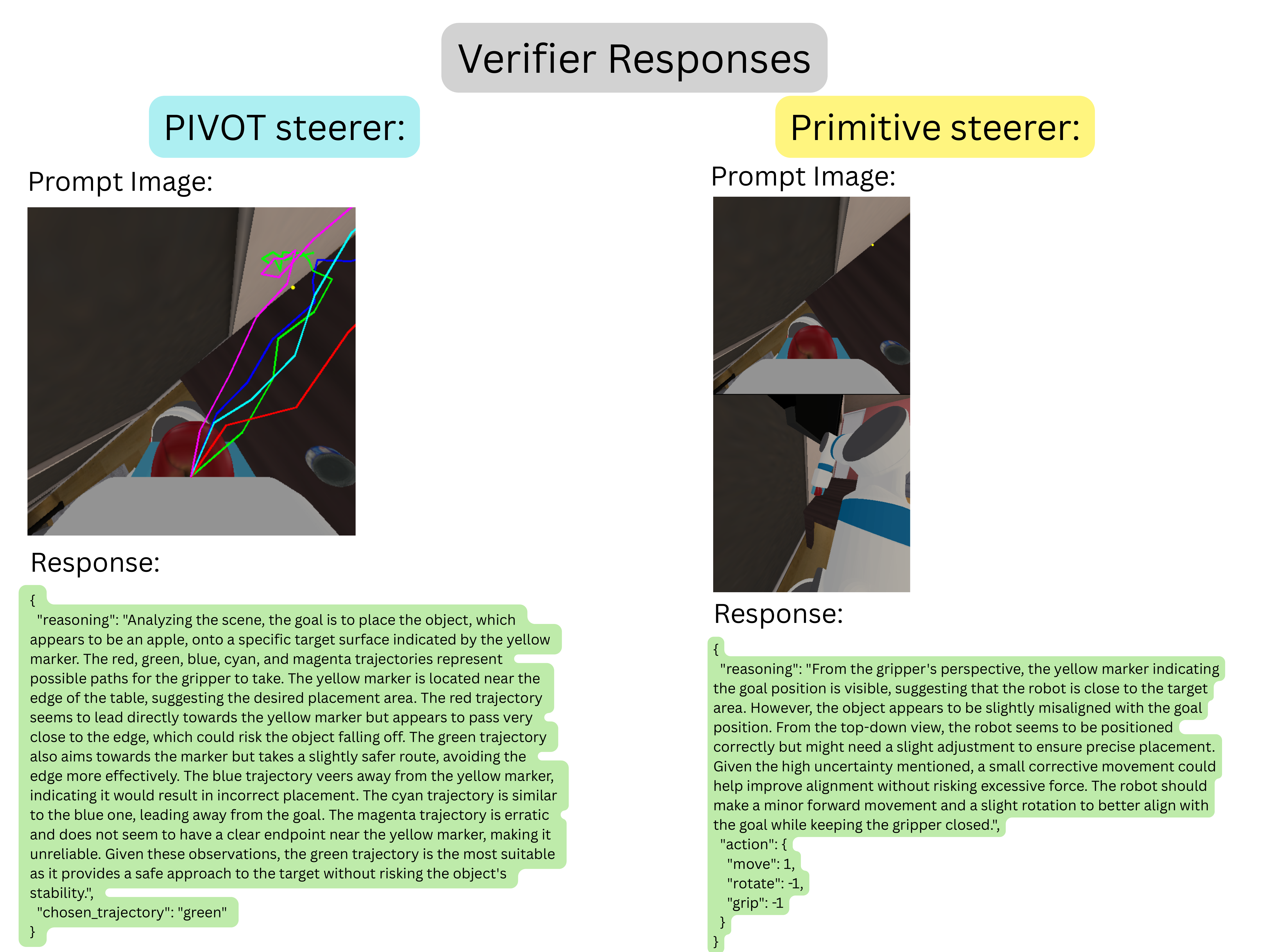}
        \caption{Verifier responses showing the PIVOT steerer trajectories and the Primitive steerer reasoning output.}
        \label{fig:verifier}
    \end{subfigure}
    
    \caption{Comparison of the steered vs unsteered rollouts and the corresponding verifier guidance. (a) Comparison of steered vs. unsteered trajectories. (b) Visualization of trajectory choices and VLM reasoning output.}
    \label{fig:steering_example}
\end{figure}

\cref{fig:steering_example} shows an example of steering with \texttt{EVE} on an instance of the \texttt{SetTable-Place} task in the MSHAB benchmark.  

\section{Hyperparameter Information and Prompts}
\label{app:hyperparameters}

\subsection{Hyperparameter Tables}

\begin{table*}[h]
\centering
\renewcommand{\arraystretch}{1.}
\setlength{\tabcolsep}{1pt}

\resizebox{0.8\textwidth}{!}{%
\begin{tabular}{@{}cc ccccc cc@{}}
\toprule
\thead{Embodiment} & \thead{MMD\\Thresh.} & \thead{Guidance\\Ratio} & \thead{Guidance\\Steps} & \thead{Sensor Img\\Res} & \thead{Num.\\Frames} & \thead{Traj.\\Perturb} & \thead{Num. Traj.\\Drawn} \\ 
\midrule
\textbf{WidowX}      & 0.9 & 40 & 4 & 640x512 & 1 & 0.01 & 5 \\
\textbf{Google Robot} & 0.8 & 40 & 2 & 640x512 & 1 & 0.01 & 5 \\
\textbf{Agile-X Dual Arm} & 1.05 & 2.5 & 15 & 1485x1182 & 1 & 0.01 & 5 \\
\bottomrule
\end{tabular}%
}
\vspace{1mm}
\caption{Hyper-parameter settings for Simpler-Env evaluations (see \cref{tab:simplerenv_results}) and Robotwin evaluations (see App. \cref{app:robotwin_result}). Traj Perturb refers to the standard deviation of gaussian noise applied to \textit{Primitive} trajectories. Num frames refers to number of frames used as history in the \textit{Primitive} steerer. Num Traj Drawn refers to number of trajectories we overlay on image for the \textit{Primitive} steerer.} 
\label{tab:hyperparams_simpler}
\end{table*}

\begin{table*}[h]
\centering

\setlength{\aboverulesep}{0pt}
\setlength{\belowrulesep}{0pt}

\renewcommand{\arraystretch}{1.4} 
\setlength{\tabcolsep}{4pt}

\resizebox{0.8\textwidth}{!}{%
\begin{tabular}{@{}ccc cccccccc@{}}
\toprule
\thead{Method} & \thead{Task} & \thead{Subtask} & \thead{MMD\\Thresh.} & \thead{Guidance\\Ratio} & \thead{Guidance\\Steps} & \thead{Sensor Img\\Res} & \thead{Num.\\Frames} & \thead{Traj.\\Perturb} & \thead{Num. Traj.\\Drawn} & \thead{Ensemble\\Ratio} \\ 
\midrule

\multirow{6}{*}{\textbf{Pivot}} 
    & \multirow{2}{*}{\makecell[l]{Prepare\\Groceries}} & Pick & 0.7 & \multirow{6}{*}{10} & \multirow{6}{*}{8} & \multirow{6}{*}{512x512} & \multirow{6}{*}{1} & \multirow{6}{*}{0.25} & \multirow{6}{*}{5} & \multirow{6}{*}{N/A} \\
    & & Place & 0.48 & & & & & & & \\
    \cmidrule(lr){2-3}
    & Tidy House & Pick & 0.7 & & & & & & & \\
    \cmidrule(lr){2-3}
    & \multirow{3}{*}{Set Table} & Pick & 0.7 & & & & & & & \\
    & & Place & 0.48 & & & & & & & \\
    & & OpenFridge & 0.48 & & & & & & & \\
\midrule

\multirow{6}{*}{\textbf{Primitive}} 
    & \multirow{2}{*}{\makecell[l]{Prepare\\Groceries}} & Pick & 0.7 & \multirow{6}{*}{10} & \multirow{6}{*}{8} & \multirow{6}{*}{512x512} & \multirow{6}{*}{1} & \multirow{6}{*}{N/A} & \multirow{6}{*}{N/A} & \multirow{6}{*}{N/A} \\
    & & Place & 0.48 & & & & & & & \\
    \cmidrule(lr){2-3}
    & Tidy House & Pick & 0.7 & & & & & & & \\
    \cmidrule(lr){2-3}
    & \multirow{3}{*}{Set Table} & Pick & 0.7 & & & & & & & \\
    & & Place & 0.48 & & & & & & & \\
    & & OpenFridge & 0.48 & & & & & & & \\
\midrule

\multirow{6}{*}{\makecell{\textbf{Pivot +}\\\textbf{Primitive}}} 
    & \multirow{2}{*}{\makecell[l]{Prepare\\Groceries}} & Pick & 0.7 & \multirow{6}{*}{10} & \multirow{6}{*}{8} & \multirow{6}{*}{512x512} & \multirow{6}{*}{1} & \multirow{6}{*}{0.25} & \multirow{6}{*}{5} & \multirow{6}{*}{1:1} \\
    & & Place & 0.48 & & & & & & & \\
    \cmidrule(lr){2-3}
    & Tidy House & Pick & 0.7 & & & & & & & \\
    \cmidrule(lr){2-3}
    & \multirow{3}{*}{Set Table} & Pick & 0.7 & & & & & & & \\
    & & Place & 0.48 & & & & & & & \\
    & & OpenFridge & 0.48 & & & & & & & \\
\bottomrule
\end{tabular}%
}
\vspace{1mm}
\caption{Hyper-parameter settings for the MSHAB evaluations (see \cref{sec:long_horizon_tasks}) Note that Guidance Ratio, Steps, Resolution, and Frames are consistent across all methods. Traj Perturb refers to standard deviation of gaussian noise added to the PIVOT trajectories before overlaying onto RGB image.}
\label{tab:hyperparams_mshab}
\end{table*}

\clearpage
\subsection{SimplerEnv Prompts}
\label{app:simplerenv_prompts}

\begin{promptbox}{Primitive Steering Prompt (Simpler-Env)}\label{prompt:primitive_simpler}
\scriptsize
You are an expert AI controller for a mobile manipulator robot in a home environment.
\tagheading{SITUATION}
\begin{itemize}[leftmargin=*]
\item The primary camera view is from the robot's overhead or wrist camera.
\item The robot has paused execution because the base policy is highly uncertain, likely due to misalignment, a potential collision, or being stuck.
\item The name of each primitive is defined below in the Primitive List.
\end{itemize}
The task is to: \texttt{<TASK\_DESCRIPTION/>}

\tagheading{SITUATION/}
\tagheading{PRIMITIVE\_LIST}
\begin{itemize}[leftmargin=*]
\item \textbf{``Nudge Left"}: translate the gripper to the left
\item \textbf{``Nudge Right"}: translate the gripper to the right
\item \textbf{``Nudge Up"}: translate the gripper vertically upwards
\item \textbf{``Nudge Down"}: translate the gripper vertically downwards
\item \textbf{``Nudge Forward"}: move the gripper forward into the scene
\item \textbf{``Retreat"}: move the gripper backward outward from the scene
\item \textbf{``Gripper Open"}: open the gripper
\item \textbf{``Gripper Close"}: close the gripper
\end{itemize}
\tagheading{PRIMITIVE\_LIST/}
Analyze the scene to diagnose the error state:
\begin{enumerate}[leftmargin=*]
\item \textbf{Misalignment:} Is the gripper too far left, right, up, or down relative to the target object?
\item \textbf{Collision:} Is the gripper pressing against a surface it shouldn't be? (Needs Retreat'' or moving backward)     \item \textbf{Air Pushing:} Is the gripper moving in free space without touching the object? (Needs Nudge Forward'')
\end{enumerate}
Select the primitive that best corrects this error to allow the robot to resume the task.
\tagheading{OUTPUT\_FORMAT}
You must conclude your response with a single, well-formed JSON object and nothing else. Do not use markdown formatting (like \texttt{```json}) or add any text before or after the JSON block.
The JSON object must contain three keys: \texttt{"reasoning"}, \texttt{"chosen\_trajectory"}, and \texttt{"gripper\_state"}.
\begin{itemize}[leftmargin=*]
\item \textbf{``reasoning"}: A string containing your detailed analysis of the error state. Explicitly mention if the gripper is misaligned (and in which direction) or if it is stuck. Explain why the chosen primitive corrects this specific error.
\item \textbf{``chosen\_trajectory"}: A string containing the name of the best primitive from the Primitive List. If no primitive helps, choose \texttt{"none"}.
\item \textbf{``gripper\_state"}: A string containing the current state of the gripper, either \texttt{"open"} or \texttt{"close"}.
\end{itemize}
Example of a perfect response format:
\begin{verbatim}
{
"reasoning": str.
"chosen_trajectory": str.
"gripper_state": str.
}
\end{verbatim}
\tagheading{OUTPUT\_FORMAT/}
\end{promptbox}
\clearpage

\begin{promptbox}{Pivot Steering (Simpler-Env)}\label{prompt:pivot_simpler}
\scriptsize
You are an expert AI controller for a mobile manipulator robot in a home environment.
\tagheading{SITUATION}
\begin{itemize}[leftmargin=*]
\item The primary camera view is from the robot's gripper. This view is overlaid with visualizations of potential future actions.
\item The image provided shows five potential future trajectories for the gripper, colored red, orange, blue, cyan, and magenta. These trajectories represent different options the robot's base policy is considering.
\item A marker at the end of each trajectory indicates the final predicted position and orientation of the gripper for that path. This marker may not always be visible.
\end{itemize}
The task is to: \texttt{<TASK\_DESCRIPTION/>}
\begin{itemize}[leftmargin=*]
\item The robot will fail the task if it encounters too high a cumulative force.
\item The robot has detected high uncertainty in its next action and requires your expert guidance to select the best path forward.
\item If all proposed trajectories appear unsafe or incorrect, it is best to reject all of them.
\end{itemize}
\tagheading{SITUATION/}

\tagheading{TRAJECTORY\_CHOICES}
\begin{itemize}[leftmargin=*]
\item \textbf{``red"}: Choose this to command the robot to follow the red path.
\item \textbf{``orange"}: Choose this to command the robot to follow the orange path.
\item \textbf{``blue"}: Choose this to command the robot to follow the blue path.
\item \textbf{``cyan"}: Choose this to command the robot to follow the cyan path.
\item \textbf{``magenta"}: Choose this to command the robot to follow the magenta path.
\item \textbf{``none"}: Choose this to reject all proposed trajectories. This is the safest option if all paths lead to failure (e.g., collision, incorrect placement).
\end{itemize}
\tagheading{TRAJECTORY\_CHOICES/}
Analyze the scene and the proposed trajectories to determine which, if any, is the best path for the robot to follow to achieve its current task.

\tagheading{OUTPUT\_FORMAT}
You must conclude your response with a single, well-formed JSON object and nothing else. Do not use markdown formatting (like \texttt{```json}) or add any text before or after the JSON block.
The JSON object must contain two keys: \texttt{"reasoning"} and \texttt{"chosen\_trajectory"}.
\begin{itemize}[leftmargin=*]
\item \textbf{``reasoning"}: A string containing your detailed analysis of the scene and each trajectory, incorporating fine-grained visual details to justify your decision. Explain why the chosen trajectory is superior and why the others are suboptimal.
\item \textbf{``chosen\_trajectory"}: A string containing one of the valid choices: \texttt{"red"}, \texttt{"orange"}, \texttt{"blue"}, \texttt{"cyan"}, \texttt{"magenta"}, or \texttt{"none"}.
\end{itemize}
Example of a perfect response format:
\begin{verbatim}
{
"reasoning": "str",
"chosen_trajectory": "str"
}
\end{verbatim}
\tagheading{OUTPUT\_FORMAT/}
\end{promptbox}
\clearpage

\subsection{Maniskill-HAB Prompts}
\label{app:mshab_prompts}

\mbox{}
\vspace{-1\baselineskip}

\begin{promptbox}{Primitive Steering Prompt (ManiSkill-HAB)}\label{prompt:primitive_mshab}
\scriptsize
You are an expert AI controller for a mobile manipulator robot in a home environment.

\tagheading{SITUATION}
\begin{itemize}[leftmargin=*]
    \item You have been given a set of images.
    \item One camera view is from the robot's gripper.
    \item The other camera view is from on top of the robot's head.
    \item If there are more than one images given for each perspective, then the images are in a sequence leading up to the current moment.
\end{itemize}
\tagheading{TASK DESCRIPTION/}
\begin{itemize}[leftmargin=*]
    \item The robot will fail the task if it encounters too high a cumulative force over the duration of the task.
    \item The robot has detected high uncertainty in its next action and may require a corrective maneuver to ensure success.
    \item If it is difficult to ascertain the correct action, it is best to not influence the policy at all (all null action).
\end{itemize}
\tagheading{SITUATION/}
\tagheading{AVAILABLE\_ACTIONS}

\textbf{Available Base Movement Actions:}
\begin{itemize}[leftmargin=3em, labelsep=1em]
    \item[\textbf{-1}:] Move the robot backwards relative to where its arm is pointing.
    \item[\textbf{0}:] Keep the robot in place; do not move.
    \item[\textbf{1}:] Move the robot forwards relative to where its arm is pointing.
    \item[\textbf{null}:] Do not influence the current action, allow the robot to continue with its current trajectory.
\end{itemize}

\vspace{0.5em}

\textbf{Available Base Rotation Actions:}
\begin{itemize}[leftmargin=3em, labelsep=1em]
    \item[\textbf{-1}:] Rotate the robot clockwise relative to a top down perspective (i.e., turn right).
    \item[\textbf{0}:] Keep the robot in place; do not rotate.
    \item[\textbf{1}:] Rotate the robot counter-clockwise relative to a top down perspective (i.e., turn left).
    \item[\textbf{null}:] Do not influence the current action, allow the robot to continue with its current trajectory.
\end{itemize}

\vspace{0.5em}

\textbf{Available Gripper Actions:}
\begin{itemize}[leftmargin=3em, labelsep=1em]
    \item[\textbf{-1}:] Continue to hold the object.
    \item[\textbf{1}:] Drop the object.
    \item[\textbf{null}:] Do not influence the current action, allow the robot to continue with its current trajectory.
\end{itemize}
\tagheading{AVAILABLE\_ACTIONS/}

Analyze the scene and determine the best action for the robot to pursue to achieve its current task.

\tagheading{OUTPUT\_FORMAT}
You must conclude your response with a single, well-formed JSON object and nothing else. Do not use markdown formatting (like \texttt{```json}) or add any text before or after the JSON block.

The JSON object must contain two keys: "reasoning" and "action".
\begin{itemize}[leftmargin=*]
    \item \textbf{"reasoning"}: A string containing your detailed analysis incorporating fine-grained visual details to justify your decisions.
    \item \textbf{"action"}: An object containing the keys "move", "rotate", and "grip", with their corresponding numerical action values.
\end{itemize}

Example of a perfect response format:
\begin{verbatim}
{ "reasoning": "The gripper is too far to the right of the goal. 
The robot needs to move its base forward and rotate slightly counter-
clockwise to align properly before attempting to place the object. 
The gripper should remain closed.",
  "action": { "move": 0, "rotate": null, "grip": -1 }}
\end{verbatim}
\tagheading{OUTPUT\_FORMAT/}

\end{promptbox}

\clearpage

\begin{promptbox}{Pivot Steering Prompt (ManiSkill-HAB)}\label{prompt:pivot_mshab}
\scriptsize
You are an expert AI controller for a mobile manipulator robot in a home environment.

\tagheading{SITUATION}
\begin{itemize}
    \item The primary camera view is from the robot's gripper. This view is overlaid with visualizations of potential future actions.
    \item The image provided shows three potential future trajectories for the gripper, colored red, green, and blue. These trajectories represent different options the robot's base policy is considering.
    \item A marker at the end of each trajectory indicates the final predicted position and orientation of the gripper for that path. This marker may not always be visible.
\end{itemize}

\tagheading{TASK DESCRIPTION/}
\begin{itemize}
    \item The robot will fail the task if it encounters too high a cumulative force.
    \item The robot has detected high uncertainty in its next action and requires your expert guidance to select the best path forward.
    \item If all proposed trajectories appear unsafe or incorrect, it is best to reject all of them.
\end{itemize}
\tagheading{SITUATION/}

\tagheading{TRAJECTORY\_CHOICES}
\begin{description}
    \item["red":] Choose this to command the robot to follow the red path.
    \item["green":] Choose this to command the robot to follow the green path.
    \item["blue":] Choose this to command the robot to follow the blue path.
    \item["none":] Choose this to reject all proposed trajectories. This is the safest option if all paths lead to failure (e.g., collision, incorrect placement).
\end{description}

Analyze the scene and the proposed trajectories to determine which, if any, is the best path for the robot to follow to achieve its current task.

\tagheading{TRAJECTORY\_CHOICES/}
\tagheading{OUTPUT\_FORMAT}
You must conclude your response with a single, well-formed JSON object and nothing else. Do not use markdown formatting (like \texttt{```json}) or add any text before or after the JSON block.

The JSON object must contain two keys: "reasoning" and "chosen\_trajectory".

\begin{itemize}
    \item \textbf{"reasoning"}: A string containing your detailed analysis of the scene and each trajectory, incorporating fine-grained visual details to justify your decision. Explain why the chosen trajectory is superior and why the others are suboptimal.
    \item \textbf{"chosen\_trajectory"}: A string containing one of the four valid choices: "red", "green", "blue", or "none".
\end{itemize}

Example of a perfect response format:
\begin{verbatim}
{
  "reasoning": "str",
  "chosen_trajectory": "str"
}
\end{verbatim}
\tagheading{OUTPUT\_FORMAT/}

\end{promptbox}

\clearpage

\subsection{RoboTwin Prompts}
\label{app:robotwin_prompts}
\begin{promptbox}{Primitive Steering Prompt (Robotwin)}\label{prompt:primitive_robotwin}
\scriptsize
You are an expert AI controller for a dual arm table-fixed manipulator robot.
\tagheading{SITUATION}
\begin{itemize}[leftmargin=*]
\item The camera view is from the robot's overhead camera.
\item While the robot has two arms, one or both arms may be off screen. If an arm comes in from the left of the image, it is the left arm and vise versa for the right.
\item The robot has paused execution because the base policy is highly uncertain, likely due to misalignment, a potential collision, or being stuck.
\item The image provided shows specific "nudge" or "retreat" actions available to correct the robot's state.
\item The name of each primitive is defined below in the Primitive List.
\item The robot will fail the task if it encounters too high a cumulative force or is too slow.
\end{itemize}
The task is to: \texttt{<TASK\_DESCRIPTION/>}
If the target object has not been picked up yet, prioritize movements which enable the robot to approach the target object from a safe angle so that it can later be grasped securely without slipping.
If the target object is already in the gripper, be careful to not destabilize it, as dropping it will take too much time to recover.
\texttt{<PRIMITIVE\_LIST>}
Analyze the scene to diagnose the error state:
\begin{enumerate}[leftmargin=*]
\item \textbf{Misalignment:} Is the gripper too far left, right, up, or down relative to the target object?
\item \textbf{Collision:} Is the gripper pressing against a surface it shouldn't be? (Needs Retreat'' or ''Nudge Back''     
\item \textbf{Air Pushing:} Is the gripper moving in free space without touching the object? (Needs Nudge Forward'')
\end{enumerate}

Select the primitive that best corrects this error to allow the robot to resume the task.
\tagheading{SITUATION/}

\tagheading{OUTPUT\_FORMAT}
You must conclude your response with a single, well-formed JSON object and nothing else. Do not use markdown formatting (like \texttt{```json}) or add any text before or after the JSON block.
The JSON object must contain three keys: \texttt{"reasoning"}, \texttt{"chosen\_trajectory\_left"}, and \texttt{"chosen\_trajectory\_right"}.
\begin{itemize}[leftmargin=*]
\item \textbf{``reasoning"}: A string containing your detailed analysis of the error state. Explicitly mention if the gripper is misaligned (and in which direction) or if it is stuck. Explain why the chosen primitive corrects this specific error.
\item \textbf{``chosen\_trajectory\_left"}: A string containing the name of the best primitive for the left arm to follow. If no primitive helps, choose \texttt{"none"}.
\item \textbf{``chosen\_trajectory\_right"}: A string containing the name of the best primitive for the left arm to follow. If no primitive helps, choose \texttt{"none"}.
\end{itemize}
Example of a perfect response format:
\begin{verbatim}
{
"reasoning": str.
"chosen_trajectory_left": str.
"chosen_trajectory_right": str.
}
\end{verbatim}
\tagheading{OUTPUT\_FORMAT/}
\end{promptbox}
\clearpage
\begin{promptbox}{Pivot Steering Prompt (Robotwin)}\label{prompt:pivot_robotwin}
\scriptsize
You are an expert AI controller for a dual arm table-fixed manipulator robot.

\tagheading{SITUATION}
\begin{itemize}
    \item The camera view is from the robot's overhead camera. This view is overlaid with visualizations of potential future actions for each arm.
    \item While the robot has two arms, one or both arms may be off screen. If an arm comes in from the left of the image, it is the left arm and vise versa for the right.
    \item The image provided shows five potential future trajectories for the gripper of each arm, colored red, orange, blue, cyan, and magenta. These trajectories represent different options the robot's base policy is considering.
    \item A marker at the end of each trajectory indicates the final predicted position and orientation of the gripper for that path. This marker may not always be visible.

    \item The task is to:  \texttt{<TASK\_DESCRIPTION/>}
    \item The robot has detected high uncertainty in its next action and requires your expert guidance to select the best path forward.
    \item If all proposed trajectories appear unsafe or incorrect, it is best to reject all of them. 
\end{itemize}
\tagheading{SITUATION/}

\tagheading{TRAJECTORY\_CHOICES}
\begin{description}
    \item["red":] Choose this to command the robot to follow the red path.
    \item["orange":] Choose this to command the robot to follow the orange path.
    \item["blue":] Choose this to command the robot to follow the blue path.
    \item["cyan":] Choose this to command the robot to follow the cyan path.
    \item["magenta":] Choose this to command the robot to follow the magenta path.
    \item["none":] Choose this to reject all proposed trajectories. This is the safest option if all paths lead to failure (e.g., collision, incorrect placement).
\end{description}
\tagheading{TRAJECTORY\_CHOICES/}
Analyze the scene and the proposed trajectories to determine which, if any, is the best path for the robot to follow to achieve its current task.

\tagheading{OUTPUT\_FORMAT}
You must conclude your response with a single, well-formed JSON object and nothing else. Do not use markdown formatting (like \texttt{```json}) or add any text before or after the JSON block.

The JSON object must contain two keys: "reasoning" and "chosen\_trajectory".

\begin{itemize}
    \item \textbf{"reasoning"}: A string containing your detailed analysis of the scene and each trajectory, incorporating fine-grained visual details to justify your decision. Explain why the chosen trajectory is superior and why the others are suboptimal.
    \item \textbf{"chosen\_trajectory\_left"}: A string containing one of the six valid choices: "red", "orange", "blue", "cyan", "magenta", or "none".
    \item \textbf{"chosen\_trajectory\_right"}: A string containing one of the six valid choices: "red", "orange", "blue", "cyan", "magenta", or "none".
\end{itemize}

Example of a perfect response format:
\begin{verbatim}
{
  "reasoning": "str",
  "chosen_trajectory\_left": "str"
  "chosen_trajectory\_right": "str"
}
\end{verbatim}
\tagheading{OUTPUT\_FORMAT/}

\end{promptbox}

\clearpage

\newcommand{\event}[1]{e_{\text{#1}}}

\newcommand{\ind}{\mathbf{1}}

\newcommand{\idx}[1]{i_{\text{#1}}}

\begin{longtable}{
    >{\raggedright\arraybackslash}p{0.22\linewidth} 
    >{\raggedright\arraybackslash}p{0.33\linewidth} 
    >{\raggedright\arraybackslash}p{0.37\linewidth}
}
    \caption{\textbf{Place Task Definitions.} $d^g_x$ denotes distance to goal. Collision threshold: 7500.} \\
    \toprule
    \multicolumn{3}{c}{\textbf{Event Definitions}} \\
    \multicolumn{3}{p{\dimexpr\linewidth-2\tabcolsep}}{\footnotesize
    $\event{grasped}: \neg \ind_{\text{g},t-1} \land \ind_{\text{g},t}$ \quad
    $\event{at goal}: d^g_{x,t-1} > 0.15 \land d^g_{x,t} \leq 0.15$ \newline
    $\event{left goal}: d^g_{x,t-1} \leq 0.15 \land d^g_{x,t} > 0.15$ \quad
    $\event{rel. at goal}: d^g_x \leq 0.15 \land \ind_{\text{g},t-1} \land \neg \ind_{\text{g},t}$ \newline
    $\event{rel. out goal}: d^g_x > 0.15 \land \ind_{\text{g},t-1} \land \neg \ind_{\text{g},t}$
    } \\
    \midrule
    \textbf{Mode} & \textbf{Description} & \textbf{Condition} \\
    \midrule
    \endfirsthead
    
    \multicolumn{3}{c}{\textit{Place Task Continued...}} \\
    \midrule
    \textbf{Mode} & \textbf{Description} & \textbf{Condition} \\
    \midrule
    \endhead

    \multicolumn{3}{l}{\textit{\textbf{Success Modes} (if $\event{success} \in E_{\text{place}}$)}} \\
    \midrule
    i. Place in goal & Releases in goal region; returns to rest. & $|E_{\text{pl}}| \leq 4 \land (\event{rel. at goal} \in E_{\text{pl}} \lor d^g_{x,0} \leq 0.15) \land \idx{left goal} \leq \idx{at goal} \land \event{excessive} \notin E_{\text{pl}}$ \\
    
    ii. Dropped to goal & Releases outside, rolls/falls in; returns to rest. & $|E_{\text{pl}}| \leq 4 \land (\event{rel. out goal} \in E_{\text{pl}} \lor d^g_{x,0} > 0.15) \land \idx{left goal} \leq \idx{at goal} \land \event{excessive} \notin E_{\text{pl}}$ \\
    
    iii. Dubious & In goal region and rest, but leaves before timeout. & $\idx{at goal} < \idx{left goal} \land \event{excessive} \notin E_{\text{pl}}$ \\
    
    iv. Winding & Leaves goal once, but eventually placed/dropped in. & $|E_{\text{pl}}| > 4 \land \idx{at goal} > \idx{left goal} \land \event{excessive} \notin E_{\text{pl}}$ \\
    
    v. Success then coll. & Success followed by excessive collisions. & $\event{excessive} \in E_{\text{pl}}$ \\
    
    \midrule
    \multicolumn{3}{l}{\textit{\textbf{Failure Modes} (if $\event{success} \notin E_{\text{place}}$)}} \\
    \midrule
    vi. Excessive coll. & Collision threshold exceeded. & $\event{excessive} \in E_{\text{pl}}$ \\
    
    vii. Didn't grasp & Fails to grasp at initialization. & $E_{\text{pl}} = () \land \event{excessive} \notin E_{\text{pl}}$ \\
    
    viii. Didn't reach & Grasps but never reaches goal region. & $|E_{\text{pl}}| > 0 \land \event{at goal} \notin E_{\text{pl}} \land \event{excessive} \notin E_{\text{pl}}$ \\
    
    ix. Place in goal fail & Placed in goal, but rolls/falls out. & $\event{at goal} \in E_{\text{pl}} \land \ind_{\text{placed latest}} \land \idx{left goal} > \idx{at goal} \land \event{excessive} \notin E_{\text{pl}}$ \\
    
    x. Dropped to goal fail & Dropped outside, rolls in, then rolls out. & $\event{at goal} \in E_{\text{pl}} \land \ind_{\text{dropped latest}} \land \idx{left goal} > \idx{at goal} \land \event{excessive} \notin E_{\text{pl}}$ \\
    
    xi. Won't let go & In goal region, but never released. & $\event{at goal} \in E_{\text{pl}} \land \idx{grasped} > \idx{rel. at goal} \land \idx{grasped} > \idx{rel. out goal} \land \event{excessive} \notin E_{\text{pl}}$ \\
    
    xii. Too slow & Released in goal, but times out before rest. & Implies $\idx{at goal} > \idx{left goal} \land \event{excessive} \notin E_{\text{pl}}$ \\
    \bottomrule
    \label{tab:place_failure_cats}
\end{longtable}

\begin{longtable}{
    >{\raggedright\arraybackslash}p{0.22\linewidth} 
    >{\raggedright\arraybackslash}p{0.35\linewidth} 
    >{\raggedright\arraybackslash}p{0.37\linewidth}
}
    \caption{\textbf{Open Task Definitions.} $a_q$ is articulation position. Collision threshold: 10000.} \\
    \toprule
    \multicolumn{3}{c}{\textbf{Event Definitions}} \\
    \multicolumn{3}{p{\dimexpr\linewidth-2\tabcolsep}}{\footnotesize
    $\event{opened}: \neg \ind_{\text{open},t-1} \land \ind_{\text{open},t}$ \quad
    $\event{closed}: \ind_{\text{open},t-1} \land \neg \ind_{\text{open},t}$ \quad
    $\event{slightly}: \neg \ind_{\text{slight},t-1} \land \ind_{\text{slight},t}$
    } \\
    \midrule
    \textbf{Mode} & \textbf{Description} & \textbf{Condition} \\
    \midrule
    \endfirsthead
    
    \multicolumn{3}{c}{\textit{Open Task Continued...}} \\
    \midrule
    \textbf{Mode} & \textbf{Description} & \textbf{Condition} \\
    \midrule
    \endhead
    
    \multicolumn{3}{l}{\textit{\textbf{Success Modes} (if $\event{success} \in E_{\text{open}}$)}} \\
    \midrule
    i. Open success & Opens and returns to rest. & $\event{excessive} \notin E_{\text{open}} \land \idx{opened} > \idx{closed}$ \\
    
    ii. Dubious & Opens, returns to rest, then closes. & $\event{excessive} \notin E_{\text{open}} \land \idx{opened} < \idx{closed}$ \\
    
    iii. Success then coll. & Opens, then excessive collisions. & $\event{excessive} \in E_{\text{open}}$ \\
    
    \midrule
    \multicolumn{3}{l}{\textit{\textbf{Failure Modes} (if $\event{success} \notin E_{\text{open}}$)}} \\
    \midrule
    iv. Excessive coll. & Collision threshold exceeded. & $\event{excessive} \in E_{\text{open}}$ \\
    
    v. Can't reach & Cannot reach articulation handle. & $\event{contact} \notin E_{\text{open}} \land \event{excessive} \notin E_{\text{open}}$ \\
    
    vi. Closed after open & Opens, but closes before rest. & $\event{closed} \in E_{\text{open}} \land \idx{closed} > \idx{opened} \land \idx{closed} > \idx{slightly} \land \event{excessive} \notin E_{\text{open}}$ \\
    
    vii. Slightly opened & Slightly opens, but not fully. & $\idx{slightly} > \idx{opened} \land \idx{slightly} > \idx{closed} \land \event{excessive} \notin E_{\text{open}}$ \\
    
    viii. Too slow & Opens, but times out before rest. & $\event{opened} \in E_{\text{open}}$ \\
    
    ix. Can't open & Reaches but cannot open. & $\event{contact} \in E_{\text{open}} \land \event{opened} \notin E_{\text{open}}$ \\
    \bottomrule
    \label{tab:open_failure_cats}
\end{longtable}

\end{document}